\documentclass[twocolumn,amsthm]{autart}    % Enable this line and disable the 
\usepackage{graphicx}
\usepackage{pgfplots}
\pgfplotsset{compat=1.18}
\usepgfplotslibrary{groupplots}
\usepackage{amsmath,etoolbox}
\usepackage{amssymb}
\usepackage{units}
\usepackage{xcolor}
\usepackage{hyperref}
\hypersetup{hidelinks}
\newcounter{part}

\usepackage[framemethod=tikz]{mdframed}
\usepackage[algo2e,ruled,vlined]{algorithm2e}
\SetKwComment{Comment}{/* }{ */}
\usepackage{subcaption}
\usepackage{natbib}
\setcitestyle{authoryear,round,colon,sort}
\usepackage{titlesec}

\titlespacing*{\section}{0pt}{14pt}{7pt}
\titlespacing*{\subsection}{0pt}{12pt}{6pt}
\makeatletter
\g@addto@macro\th@plain{%
  \thm@preskip=6pt
  \thm@postskip=6pt
}
\g@addto@macro\th@definition{%
  \thm@preskip=6pt
  \thm@postskip=6pt
}
\makeatother

\graphicspath{{Figures/}}
\newcommand{\mat}[1]{\boldsymbol{#1}}
\renewcommand{\vec}[1]{\boldsymbol{#1}}

\newcommand{\Real}{\mathbb{R}}

\newcommand{\mr}[1]{\mathrm{#1}}
\newcommand{\T}{^{\mathop{\mathrm{T}}}}

\makeatletter
\preto\env@matrix{}
\makeatother

\newmdenv[
    frametitlerule=true,
    frametitlebackgroundcolor=gray!17,
    frametitlefont=\small\bfseries,
    backgroundcolor=gray!7,
    outerlinewidth=2pt,
    linecolor=gray!70,
    roundcorner=15pt,
    innerleftmargin=0pt,
    innerrightmargin=4pt
]{infobox}

\theoremstyle{plain}
\newtheorem{assumption}{Assumption}
\newtheorem{proposition}{Proposition}
\newtheorem{lemma}{Lemma}
\theoremstyle{definition}
\newtheorem{definition}{Definition}
\newtheorem{remark}{Remark}

\begin{document}

\onecolumn
\thispagestyle{empty}

\vspace*{6mm}

\begin{center}
    This preprint was submitted for publication in \textit{Automatica}, the Journal of the International Federation of Automatic Control (IFAC).\\[3mm]
    First submission: 28 October 2026\\[3mm]
    \href{https://www.sciencedirect.com/journal/automatica}
         {https://www.sciencedirect.com/journal/automatica}\\[15mm]

    An earlier version of this manuscript was submitted to
    \textit{The International Journal of Robotics Research (IJRR)}.
    The submission history is summarized below.\\[3mm]

    Major revision requested and notification of scope mismatch: 18 August 2026\\[3mm]
    Revised manuscript submitted: 31 March 2026\\[3mm]
    Major revision requested: 06 April 2025\\[3mm]
    First submission: 12 October 2024\\[3mm]
    \href{https://journals.sagepub.com/home/ijr}
         {https://journals.sagepub.com/home/ijr}
\end{center}

\vspace*{\fill}

\clearpage
\twocolumn

% Main paper starts here

\begin{frontmatter}

\title{Indirect Optimal Control for Constructing Libraries of Periodic Trajectories: Application to Bipedal Walking} % Title, preferably not more 

\author[stuttgart]{Maximilian Raff}\ead{raff@iams.uni-stuttgart.de},
\author[saarland]{Kathrin Fla{\ss}kamp},
\author[stuttgart]{C. David Remy}

\address[stuttgart]{Institute for Adaptive Mechanical Systems, University of Stuttgart, Heisenbergstr. 3, 70569 Stuttgart, Germany}
\address[saarland]{Systems Modeling and Simulation, Saarland University, Campus A 5 1, 66123 Saarbr\"ucken, Germany}

                             % keyword list or with the 
                                          % help of the Automatica 
                                          % keyword wizard

\begin{abstract}                          % Abstract of not more than 200 words.
Trajectory optimization is a fundamental tool for generating efficient, dynamically consistent periodic motions, commonly referred to as gaits, in legged locomotion.
This paper explores the indirect method of periodic trajectory optimization, emphasizing its application in creating optimal gaits for legged systems and contrasting it with the more common direct method.
While the direct method offers implementation flexibility, it is limited by its need for an input space parameterization.
In contrast, the indirect method achieves higher accuracy by computing the control input from states and costates along the optimal trajectory.
In this work, we tackle the convergence challenges associated with indirect shooting methods by utilizing numerical continuation methods.
This is particularly useful for the systematic development of gait libraries.
Our contributions include: 
(1) the formalization of a general periodic trajectory optimization problem that extends existing first-order necessary conditions to a broader range of cost functions and operating conditions; 
(2) a methodology for efficiently generating optimal gait libraries using single shooting and numerical continuation;
(3) a novel approach for reconstructing Lagrange multipliers and costates from passive gaits; 
(4) a comparative analysis of the indirect and direct shooting methods using a compass-gait walker as a case study, demonstrating the higher accuracy of the indirect method; and
(5) demonstrating applicability to the more complex legged robot RABBIT, with ten dynamic states and four inputs.
These results establish a systematic indirect optimal control framework for synthesizing families of optimal periodic trajectories in legged locomotion.
\end{abstract}

\end{frontmatter}

\section{Introduction}
\label{sec:intro}
Optimal control is widely used for path and motion planning in robotics, particularly in legged systems, where efficient, stable, and dynamically consistent motions are essential.
% Optimal control is widely used for path and motion planning in robotics.
% This holds in particular in the field of legged robotic systems, where the creation of efficient, stable, and dynamically consistent gaits is a crucial ability.
Trajectory optimization has thus become an essential tool for design and control in applications that reach from humanoid robots to exoskeletons and prosthetics ~\citep{Wensing2024}.

Methods for optimal control are classified into \emph{direct} and \emph{indirect}. The \emph{direct method} is the most commonly used for trajectory optimization.
In this method, control inputs are parameterized and optimized directly over a predefined input space~\citep{Diehl2006,Kelly2017,Wensing2024}.
This discretization typically leads to a high-dimensional nonlinear program but offers conceptual simplicity and considerable implementation flexibility.
As a drawback, the accuracy of the resulting optimal trajectory depends on the chosen parameterization of the input space, potentially missing better solutions that exist outside this discretization.
This leads to a trade-off between accuracy and problem size, while the true global optimum may remain unattainable due to the constraints imposed by the parameterization~\citep{Houska2013,Lin2014}.
The \emph{indirect method} offers a different approach by implicitly defining the control inputs as functions of the system's states and additional costates. 
It often provides more accurate solutions, especially when using shooting methods with variable step-size integrators.
While the introduction of costates helps formalize the first-order necessary conditions, it does not significantly increase the dimensionality of the optimization problem.
The challenge of the indirect method, however, lies in the numerical convergence, primarily due to the conservative dynamics of the Hamiltonian system~\citep{Gros2022}. 
Specifically, the coexistence of stable and unstable modes can make shooting methods highly sensitive to initial costate estimates, leading to convergence difficulties~\citep{Stryk1992,Betts2010,Gros2022,Wensing2024}.

Although the indirect method in periodic trajectory optimization has been applies in fields such as aerospace~\citep{Gilbert1981, Speyer1996} and hybrid systems in chemical processes~\citep{Horn1967}, it remains less popular than the direct method.
An example from legged locomotion is the work by \cite{Channon1996} on bipedal walking.
However, this work utilized collocation techniques, a common approach for extending the convergence region of the indirect method~\citep{Stryk1992}.
Unfortunately, combining the indirect method with collocation or fixed step-size shooting diminishes its core advantage, making it nearly equivalent to the direct method in terms of accuracy~\citep{Ross2020}, and thus losing its potential for achieving higher precision in solving optimal control problems.
In addition, \cite{Channon1996} did not consider period-normalized cost functions, which are commonly used in periodic trajectory optimization~\citep{Colonius1988, Speyer1984, Horn1967} and essential for evaluating key performance metrics such as the cost of transport.

A key insight of this work is that in legged robotics, we are often interested not only in a single solution but in generating libraries of optimal gaits, i.e., periodic motions, that adapt to varying conditions, such as changes in terrain or the robot's speed \citep{Liu2009,Liu2012,Rosa2022}.
When considering such parameterized optimization problems, numerical continuation methods can quickly generate many gaits.
These methods offer robust local convergence properties and can help overcome the challenges associated with indirect methods. Combined with a single shooting approach, they hold promise for generating families of optimal gaits with higher accuracy than direct methods. 
In prior work, we showed that such parameterized optimization problems may include \textit{passive gaits}, which are especially relevant as they trivially solve certain economic optimization problems by minimizing energy consumption without requiring active control inputs \citep{Raff2022b,Rosa2023}.
Such passive gaits could be leveraged as seeds for a gait library, if we can reconstruct states, costates, and Lagrange multipliers as they are needed in the indirect method.

This paper explores these ideas and makes several contributions to trajectory optimization for systems with hybrid dynamics, particularly in legged locomotion:

\begin{itemize}
    \item \textbf{Formalization of a General Trajectory Optimization Problem}: We extend the first-order necessary conditions from the existing literature to a broader class of cost functions and operating conditions, addressing both legged systems and other periodically operating robots with hybrid dynamics.

    \item \textbf{Generation of Gait Libraries}: 
    By reformulating the first-order necessary conditions as parameter-dependent root-finding problems and applying a single shooting method, we develop a robust framework to generate and analyze families of optimal gaits using numerical continuation techniques.

    \item \textbf{Lagrange Multiplier and Costate Reconstruction from Passive Gaits}: 
    We present a systematic methodology for reconstructing costates based on observability arguments.
    
    \item \textbf{Performance Evaluation of the Indirect Method}: 
    Using the compass-gait walker as a case study, we compare the indirect with the direct shooting method. 
    The results highlight the superior accuracy of the indirect method and its advantages in handling input space parameterization for optimal gait generation in hybrid dynamical systems.
    %The results highlight the indirect method’s superior accuracy and its effectiveness in managing input space parameterization, showcasing its advantages for generating optimal gaits in systems with hybrid dynamics.

    \item \textbf{Application to Complex Legged Systems}: 
    To illustrate the method's applicability to more complex robots, we apply the framework to the RABBIT robot, with a 10 states and four control inputs.
    
\end{itemize}

The remainder of this paper is organized as follows. 
In Section~\ref{sec:theory}, we formalize a generic trajectory optimization problem applicable to legged systems and other periodically operating robots. 
We derive the first-order necessary conditions using the Calculus of Variations and the Lagrange multiplier method, extend the framework to parameterized families of optimal gaits, and discuss multiplier reconstruction from passive periodic solutions.
%This section includes a detailed derivation of the first-order necessary conditions for optimality using the Calculus of Variations and the Lagrange multiplier method. 
%We then extend the optimization framework to incorporate a set of parameters that characterize families of optimal gaits, concluding with a discussion on how to reconstruct Lagrange multipliers from passive periodic solutions within these gait families.
In Section~\ref{sec:implementation}, we discuss the implementation of single-shooting methods and numerical continuation techniques for tracing solution curves within the parameterized optimization problem.
Section~\ref{sec:example} presents the compass-gait walker as a case study, applying the indirect shooting method to bipedal walking. This section demonstrates how to generate families of optimal gaits that originate from passive gaits, and compares the indirect method to direct shooting approaches.
Similarly, Section~\ref{sec:RABBIT} applies the same procedure to the more complex legged robot RABBIT, highlighting the indirect method’s applicability to more complex systems.
Section~\ref{sec:disc} concludes the paper and is followed by an appendix that provides the derivation of the first-order optimality conditions (Appendix~\ref{sec:appendixVariation}), the predictor-corrector continuation method (Appendix~\ref{sec:appendixContinuation}), the dynamics of the compass-gait walker (Appendix~\ref{sec:cgw}), and additional details on the implementation of the direct shooting method (Appendix~\ref{sec:appendixDirect}).

\section{Theory}
\label{sec:theory}
In this section, we formalize a generic trajectory optimization problem applicable to legged systems as well as other periodically operating robots and mechanisms that may involve non-smooth interactions with the environment. 
We present a detailed derivation of the first-order necessary conditions using the Calculus of Variations, incorporating the Lagrange multiplier method. 
The section concludes with a parameterization for a library of optimal trajectories and a discussion on reconstructing Lagrange multipliers from passive periodic solutions.

To avoid notational overhead, we omit the argument of a function when it is clear from the context, writing $\alpha$ instead of $\alpha(t)$.
Otherwise, we use the shorthand notation $\alpha_t=\alpha(t)$.

\subsection{Problem Formulation}
Periodic trajectory optimization seeks a period~$T\in\Real$, a state trajectory~$\vec{x}(\cdot)$, and an input trajectory~$\vec{u}(\cdot)$, with $\vec{x}(t)\in\Real^{n_\mr{x}}$ and $\vec{u}(t)\in\Real^{n_\mr{u}}$ for $t\in[0,T]$, that (locally) minimize the optimization problem~$\mathcal{P}$ (see box below).
\begin{figure}[h]
    \centering
    \begin{infobox}[frametitle={~~~Periodic Trajectory Optimization Problem~$\mathcal{P}$}]
    \vspace{-10pt}
\begin{subequations}\label{eq:indirectOptProblem}
\begin{alignat}{2}
& \underset{T, \vec{x}(\cdot), \vec{u}(\cdot)}{\text{minimize}}
  & \quad & c(T, \vec{x}(T), y(T)) \label{eq:indMayerCost} \\
& \text{subject to} & & \dot{\vec{x}}(t) = \vec{f}\big(\vec{x}(t), \vec{u}(t)\big),\quad t\in[0,T],\label{eq:indDynOpt1}\\
& & & \vec{x}(0) = \vec{g}(\vec{x}(T)),\label{eq:boundCond1} \\[2mm]
& & &      \dot{y}(t) = l\big(\vec{x}(t), \vec{u}(t)\big),\quad t\in[0,T],\label{eq:indDynOpt2}\\
& & &       y(0) = 0, \label{eq:boundCond2} \\[2mm]
& & & \underbrace{\begin{bmatrix}
    e\big(T, \vec{x}(T), \vec{x}(0)\big)  \\
    \vec{\omega}\big(T, \vec{x}(T), \vec{x}(0)\big) 
\end{bmatrix}}_{=\vec{h}\big(T, \vec{x}(T), \vec{x}(0)\big) } = \vec{0}. \label{eq:indConst}
\end{alignat}
\end{subequations}
\end{infobox}
\end{figure}

The problem~$\mathcal{P}$ is formulated in general Mayer form~\citep{Liberzon2011}, with terminal cost function~$c:\Real\times\Real^{n_\mr{x}}\times\Real\to\Real$. 
The auxiliary state~$y(\cdot)$, with~$y(t)\in\mathbb{R}$, accumulates the instantaneous cost defined by~$l:\mathbb{R}^{n_\mr{x}}\times\mathbb{R}^{n_\mr{u}}\to\mathbb{R}$, such that~$y(T)$ represents the total accumulated cost and enters the Mayer-type cost function.
% The auxiliary state~$y(\cdot)$, with~$y(t) \in \mathbb{R}$, accumulates the instantaneous~cost over time through the function~$l: \mathbb{R}^{n_\mr{x}} \times \mathbb{R}^{n_\mr{u}} \to \mathbb{R}$. Consequently,~$y(T)$ represents the total accumulated instantaneous cost, which can then be incorporated into the Mayer-type cost function.
Additionally, $\mathcal{P}$ includes dynamic constraints that represent a simple hybrid system~\citep{Ames2006}, consisting of continuous and discrete dynamics in a single phase.
The continuous evolution is governed by the ordinary differential~\eqref{eq:indDynOpt1} with vector field~$\vec{f}: \Real^{n_\mr{x}}\times\Real^{n_\mr{u}}\to \mathbb{R}^{n_\mr{x}}$, while the discrete transitions are modeled by the reset map~$\vec{g}:\mathbb{R}^{n_\mr{x}}\to\mathbb{R}^{n_\mr{x}}$ in~\eqref{eq:boundCond1}, which also enforces periodicity. 
The guard condition associated with these discrete transitions is provided by the event constraint~$e:\Real\times\Real^{n_\mr{x}}\times\Real^{n_\mr{x}}\to\Real$. 
This event typically depends on the final state~$\vec{x}(T)$ and time~$T$, but may also involve the initial state~$\vec{x}(0)$ due to periodicity as in \cite{Channon1996}.

The equality constraint~$\vec{h}:\Real\times\Real^{n_\mr{x}}\times\Real^{n_\mr{x}}\to\Real^{1+n_{\omega}}$ in~\eqref{eq:indConst} encompasses both the event constraint~$e$ and an operating point constraint~$\vec{\omega}:\Real\times\Real^{n_\mr{x}}\times\Real^{n_\mr{x}}\to\Real^{n_{\omega}}$, which imposes $n_{\omega}$ operating conditions.
These typically specify desired amplitudes, frequencies, or metrics such as average energy or velocity.
The operating point prescribes motion characteristics while excluding trivial optima at equilibria or zero duration~($T=0$), particularly when optimizing energetic efficiency.

\begin{remark}
    By adopting a hybrid dynamical framework, non-smooth interactions with the environment are assumed to be known \emph{a priori}, allowing the dynamics to be separated into continuous ($\vec{f}$) and discrete ($\vec{g}$) components. Furthermore, incorporating the discrete dynamics into the boundary conditions of the periodic trajectory optimization problem~$\mathcal{P}$ renders the formulation differentiable under the following assumptions.
\end{remark}

\begin{assumption}[Existence of Local Minimizer]\label{assumption:existence}
The optimization problem~$\mathcal{P}$ admits at least one strict local minimizer~$(T^\ast>0,\vec x^\ast(\cdot),\vec{u}^\ast(\cdot))$.
\end{assumption}
\begin{assumption}[Continuous Differentiability]\label{assumption:continuous}
The control input is continuous, i.e., $\vec u(\cdot) \in C^0$, 
and unconstrained. The functions $c(\cdot,\cdot,\cdot)$, $\vec f(\cdot,\cdot)$, $l(\cdot,\cdot)$,  $\vec{g}(\cdot)$ and $\vec{h}(\cdot,\cdot,\cdot)$ are of class $C^1$ with respect to all arguments.
\end{assumption}
In the absence of control constraints, continuity of the optimal input generally follows from the smoothness of both the system dynamics and the cost functions. Therefore, this choice is not expected to result in any performance degradation.
More generally, the continuity assumption on the control input $\vec{u}$ can be relaxed 
(e.g., \citep{Speyer1984,Liberzon2011}) by invoking Pontryagin's Maximum Principle \citep{Pontryagin1987}. However, such generalizations are beyond the scope of this paper.
Additionally, we restrict ourselves to equality constraints in $\mathcal{P}$. 
A potential extension to include inequality constraints on the states~$\vec{x}$ and inputs~$\vec{u}$ is discussed in the concluding Section~\ref{sec:disc}.

\subsection{Derivation of the First-Order Necessary Condition}\label{sec:necessaryConditionsDerivation}
To formalize the first-order necessary conditions of problem~$\mathcal{P}$, we utilize Lagrange's multiplier method for the dynamics~\eqref{eq:indDynOpt1}, \eqref{eq:indDynOpt2} and boundary constraints~\eqref{eq:boundCond1}, \eqref{eq:boundCond2}, \eqref{eq:indConst} \citep{Liberzon2011}. 
The dynamics~\eqref{eq:indDynOpt1} and \eqref{eq:indDynOpt2} must hold point-wise in time for all $t\in[0,T]$.
They introduce continuous time Lagrange multipliers, referred to as the costates~$\vec p(\cdot)$, with $\vec p(t)\in\Real^{n_\mr{x}}$ and $q(\cdot)$, with $q(t)\in\Real$, of $\vec x$ and~$y$, respectively.
The constraints~\eqref{eq:indConst} also introduce Lagrange multipliers~$\vec \lambda\T=[\lambda_\mr{e}~\vec{\lambda}_\omega\T]\in\Real^{1+n_\omega}$.
However, these multipliers are discrete in time, as they act at constraints that depend only on the states at the boundary of the trajectory.
Similarly, for \eqref{eq:boundCond1} and \eqref{eq:boundCond2}, we introduce Lagrange multipliers~${}_\mr{z}\vec{\lambda}\T=[{}_\mr{x}\vec{\lambda}\T~{}_\mr{y}\lambda]$, where ${}_\mr{x}\vec{\lambda}\in\Real^{n_\mr{x}}$ and ${}_\mr{y}\lambda\in\Real$.

For notational convenience, we group the continuous and discrete time variables as
\begin{equation}\label{eq:vwzDef}
\left.\begin{aligned}
    \vec z\T &= \begin{bmatrix}
        \vec x\T & y
    \end{bmatrix}\\
    \vec \rho\T &= \begin{bmatrix}\vec p\T & q\end{bmatrix}
\end{aligned}~\right\}~
\vec w(\cdot)=\begin{bmatrix}\vec z(\cdot)\\\vec\rho(\cdot)\\\vec u(\cdot)\end{bmatrix},\quad
\text{and~}
\vec v=\begin{bmatrix}T\\\vec\lambda\end{bmatrix}.
\end{equation}
% \begin{subequations}\label{eq:vwzDef}
% \begin{align}
%     \vec z(\cdot)\T &= \begin{bmatrix}
%         \vec x(\cdot)\T & y(\cdot)
%     \end{bmatrix},\\
%     \vec \rho(\cdot)\T &= \begin{bmatrix}\vec p(\cdot)\T & q(\cdot)\end{bmatrix},\\
%     \vec w(\cdot)\T &= \begin{bmatrix}\vec z(\cdot)\T & \vec \rho(\cdot)\T & \vec u(\cdot)\T\end{bmatrix},\\
%     \vec v\T &=\begin{bmatrix}T & \vec \lambda\T\end{bmatrix},
% \end{align}
% \end{subequations}
and define the Hamiltonian as
\begin{equation}\label{eq:hamiltonian}
        \mathcal{H}(\underbrace{\vec x,y,\vec p,q,\vec u}_{=\vec{w}}) = \vec p\T \vec f(\vec x,\vec u)+q\,l(\vec x,\vec u).
\end{equation}

With the introduction of these Lagrange multipliers, we follow the classical idea of Joseph-Louis Lagrange (cf. Section 2.5 in \cite{Liberzon2011}) and incorporate both the dynamics and boundary constraints of~$\mathcal{P}$ into the terminal cost~$c$, yielding the augmented Lagrangian:
\begin{subequations}\label{eq:lagrangian}
\begin{equation}
    L\big({}_\mr{z}\vec{\lambda},\vec{v},\vec{w}\big) = L_1(\vec{v},\vec{w})+L_2({}_\mr{z}\vec{\lambda},\vec{w})+ L_3(\vec{v},\vec{w}),
\end{equation}
where
\begin{align}
    L_1&=c\big(T,\vec z(T)\big)+ \vec\lambda\T\vec h\big(T,\vec x(T),\vec{x}(0)\big), \\
    L_2&={}_\mr{x}\vec{\lambda}\T\big(\vec{g}(\vec{x}(T))-\vec{x}(0)\big)+{}_\mr{y}\lambda \,y(0), \\
    L_3&= \int\displaylimits_0^T \mathcal{H}(\vec w)-\vec \rho\T \dot{\vec z}\mr{d}t,
\end{align}
\end{subequations}
With the Lagrangian, the stationarity condition of the constrained optimization problem~$\mathcal{P}$ can be equivalently written as an unconstrained problem~\citep{Kalman2009}:
\begin{equation}\label{eq:problemStatement}
    \underset{{}_\mr{z}\vec{\lambda}, \vec v,\vec w(\cdot)}{\text{stationary}}
  \quad L\big({}_\mr{z}\vec{\lambda},\vec{v},\vec{w}(\cdot)\big).
\end{equation}
\begin{remark}\label{remark:Regular}
    Formulating the unconstrained problem~\eqref{eq:problemStatement} using Lagrange multipliers requires regularity conditions on the equality constraints. Their technical treatment in infinite-dimensional vector spaces is beyond the present scope; see Chapter~9.2 of~\cite{Luenberger1997} for details.
    % To formulate the unconstrained problem~\eqref{eq:problemStatement} using Lagrange multipliers, certain regularity conditions on the equality constraints are required. 
    % We have not introduced the notion of regularity here, as it involves technical details specific to the infinite-dimensional vector spaces under consideration. 
    % For a thorough discussion on the required regularity conditions, please refer to Chapter~9.2 of \cite{Luenberger1997}. 
    
   % It is also important to emphasize that the unconstrained problem~\eqref{eq:problemStatement} is intended solely for deriving first-order necessary conditions. Its second-order variation would always reveal saddle points~\citep{Kalman2009} and should not be interpreted as a minimization problem or as an equivalent reformulation of the original problem~$\mathcal{P}$.
\end{remark}
\begin{proposition}[First-Order Optimality]\label{prop:firstOrderNecessary}
Under Assumptions~\ref{assumption:existence} and \ref{assumption:continuous}, at a strict local minimizer of~$\mathcal{P}$, there exist costates~$(\vec{p},q)$ and Lagrange multipliers $\vec{\lambda}$ that satisfy the differential-algebraic system of~(DAEs)~\eqref{eq:IndirectCond1} with boundary and transversality conditions (BC/TC)~\eqref{eq:IndirectCond2} (see box below).

\begin{figure}[h]
    \centering
    \begin{infobox}[frametitle={~~~First-Order Necessary Condition of Optimality}]
    \begin{subequations}\label{eq:IndirectCond1}
    ~DAE:
    \begin{alignat}{2}
    \dot{\vec{x}} &= \frac{\partial \mathcal{H}}{\partial \vec p}\T &&= \vec f(\vec x,\vec u),\\
        \dot{y} &= \frac{\partial \mathcal{H}}{\partial q} &&= l(\vec x,\vec u),\\
        \dot{\vec{p}} &= -\frac{\partial \mathcal{H}}{\partial \vec x}\T &&=- \frac{\partial \vec{f}}{\partial \vec{x}}\T \vec p- \frac{\partial l}{\partial \vec{x}}\T q,\\
        q &\equiv \frac{\partial c}{\partial y_T},\label{eq:qValue}\\
        \vec{0} &= \frac{\partial \mathcal{H}}{\partial \vec u}\T &&= \frac{\partial \vec{f}}{\partial \vec{u}}\T \vec p+ \frac{\partial l}{\partial \vec{u}}\T q.
    \end{alignat}
    \end{subequations}
    ~BC/TC:
    \begin{subequations}\label{eq:IndirectCond2}
    \begin{align}
    \vec{g}\left(\vec{x}(T)\right)-\vec x(0)&=\vec{0},\\
    y(0)&=0,\\
       \vec{h}\big(T,\vec x(T),\vec{x}(0)\big) &=\vec{0},\\
       \mathcal{H}\big(\vec w(T)\big)+\frac{\partial c}{\partial T}+\vec{\lambda}\T\frac{\partial \vec{h}}{\partial T}&=0,\label{eq:trans_t}\\
       \frac{\partial \vec{g}}{\partial \vec x_T}\T\cdot\left(\vec{p}(0)+\frac{\partial \vec{h}}{\partial \vec{x}_0}\T\vec{\lambda}\right)-\vec{p}(T)&\notag\\
       + \frac{\partial c}{\partial \vec x_T}\T+\frac{\partial \vec{h}}{\partial \vec x_T}\T\vec{\lambda}&=\vec{0}.\label{eq:trans_x}
    \end{align}
    \end{subequations}
    \end{infobox}
\end{figure}
\end{proposition}

\begin{pf}
\eqref{eq:IndirectCond1} and \eqref{eq:IndirectCond2} are derived using the Calculus of Variations, as detailed in Appendix~\ref{sec:appendixVariation}.
\end{pf}

\subsection{Parameterized Optimization Problem}
In legged locomotion, there is a significant interest in generating libraries of gaits that adapt to varying conditions which are represented by a parameter set $\vec{\sigma}\in\Real^{n_\sigma}$~\citep{Liu2012,Reher2021,Westervelt2007,Raff2022b,Rosa2023}. 
These parameters may represent environmental variations, such as changes in slope or terrain, as well as variations in the robot's operating conditions, such as forward speed or step length.
Introducing these parameters in the optimization problem~$\mathcal{P}$ creates the parameterized optimization problem:
\begin{equation*}
\mathcal{P}_{\vec{\sigma}}\left\{
    \begin{array}{ll}
    \underset{T, \vec{x}(\cdot), \vec{u}(\cdot)}{\text{minimize}}
     & c(T, \vec{x}(T), y(T);\vec{\sigma})  \\
    \text{subject to} & \dot{\vec{x}}(t) = \vec{f}\big(\vec{x}(t), \vec{u}(t);\vec{\sigma}\big),\quad t\in[0,T],\\
    & \vec{x}(0) = \vec{g}(\vec{x}(T);\vec{\sigma}), \\[2mm]
    &      \dot{y}(t) = l\big(\vec{x}(t), \vec{u}(t);\vec{\sigma}\big),\quad t\in[0,T],\\
    &       y(0) = 0, \\[2mm]
    & \vec{h}\big(T, \vec{x}(T), \vec{x}(0);\vec{\sigma}\big) = \vec{0},
    \end{array}\right.
\end{equation*}
where all functions are defined analogously to~\eqref{eq:indirectOptProblem}.
With the definition of $\mathcal{P}_{\vec{\sigma}}$, the first-order optimality conditions~\eqref{eq:IndirectCond1}, \eqref{eq:IndirectCond2} are also parameterized by $\vec{\sigma}$.
Provided that a stationary point of~$\mathcal{P}_{\bar{\vec{\sigma}}}$ is regular (Remark~\ref{remark:Regular}) and differentiable with respect to a set of nominal parameters~$\bar{\vec{\sigma}}$, there exists a locally defined manifold of optimal solutions parameterized by $\vec{\sigma}$.
\subsection{Reconstruction of Lagrange Multipliers or Passive-Optimal Motions}\label{sec:reconstructLagrange}
Unlike the Lagrange multipliers ${}_\mr{z}\vec{\lambda}$ in \eqref{eq:lamxy}, the multipliers $\vec{\lambda}$ and $\vec{\rho}(\cdot)$ cannot be eliminated as easily.
In particular, determining the costates~$\vec{p}(\cdot)$ is challenging in indirect methods, as discussed, for example, by \cite{Stryk1992} and in Chapter 4.4 of \cite{Betts2010}.

In this paper, we exploit \emph{passive-optimal solutions}, defined as locally optimal solutions of the parameterized problem~$\mathcal{P}_{\vec{\sigma}}$ at a fixed parameter value~$\bar{\vec{\sigma}}$ with vanishing control inputs~($\vec{u}\equiv\vec{0}$).
For such solutions, the associated costates can be reconstructed.
If they lie on the solution manifold (i.e., the library of optimal motions) described above, they provide suitable initial conditions for a numerical continuation procedure, enabling a systematic exploration of the entire manifold.

\begin{definition}[Passive-Optimal Solution] A strict local minimizer $(T^\ast,\vec x^\ast(\cdot),\vec{u}^\ast(\cdot))$, with $\vec{u}^\ast\equiv \vec{0}$, at a fixed parameter vector~$\bar{\vec{\sigma}}$ for problem~$\mathcal{P}_{\bar{\vec{\sigma}}}$, is referred to as a passive-optimal solution, or in the context of legged locomotion, a passive-optimal gait.
\end{definition}

To apply the reconstruction technique for the multipliers in a passive-optimal solution, we reformulate the necessary conditions~\eqref{eq:IndirectCond1} to
\begin{subequations}\label{eq:observeDyn}
\begin{align}
    \dot{\vec{w}}&=\vec{\digamma}(\vec{w},\bar{\vec{\sigma}}),\\
    \vec{0} &= \mathcal{H}_u(\vec{w},\bar{\vec{\sigma}}),\label{eq:H_u}
\end{align}
\end{subequations}
where
\begin{subequations}
\begin{align}
    \vec{\digamma}&:=\begin{bmatrix}
    \vec{f}(\vec{x},\vec{u},\bar{\vec{\sigma}})\\
    l(\vec{x},\vec{u},\bar{\vec{\sigma}})\\
    - \dfrac{\partial \vec{f}}{\partial \vec{x}}\bigg\vert_{(\vec{x},\vec{u},\bar{\vec{\sigma}})}\T \cdot\vec p- \dfrac{\partial l}{\partial \vec{x}}\bigg\vert_{(\vec{x},\vec{u},\bar{\vec{\sigma}})}\T \cdot q\\
    0\\
    \vec{0}
    \end{bmatrix},\label{eq:F}\\
    \mathcal{H}_u &:= \frac{\partial \vec{f}}{\partial \vec{u}}\bigg\vert_{(\vec{x},\vec{u},\bar{\vec{\sigma}})}\T\cdot \vec{p}+\frac{\partial l}{\partial \vec{u}}\bigg\vert_{(\vec{x},\vec{u},\bar{\vec{\sigma}})}\T\cdot q.
\end{align}
\end{subequations}
Note that the dynamics~$\dot{\vec{u}}=\vec{0}$ in \eqref{eq:F} result from the passive inputs~$\vec{u}^\ast\equiv \vec{0}$.
\begin{assumption}[Smooth Dynamics]\label{assumption:smooth}
At the fixed parameter vector~$\bar{\vec{\sigma}}$, the dynamics~$\vec{f}$ and instantaneous cost~$l$ are of class~$C^\infty$ with respect to~$\vec{x}$ and~$\vec{u}$.
\end{assumption}
\begin{assumption}[Unique Multipliers]\label{assumption:uniqueness}
At each strict local minimizer of~$\mathcal{P}_{\bar{\sigma}}$ under consideration, the costates $(\vec{p}(\cdot),q)$ and Lagrange multipliers~$\vec{\lambda}$ satisfying~\eqref{eq:IndirectCond1} and~\eqref{eq:IndirectCond2} are unique under suitable regularity conditions on the equality constraints (cf. Remark~\ref{remark:Regular}).
\end{assumption}

\begin{lemma}[Obeservability of the Costates]\label{lem:observe}
Under Assumptions~\ref{assumption:smooth} and~\ref{assumption:uniqueness}, consider a passive-optimal solution $(T^\ast,\vec{x}^\ast(\cdot),\vec{u}^\ast(\cdot))$ of~$\mathcal{P}_{\bar{\vec{\sigma}}}$ with~$\vec{u}^\ast\equiv\vec{0}$.
The associated costates~$(\vec{p}(\cdot),q)$ can be reconstructed from the first-order necessary conditions~\eqref{eq:IndirectCond1} and \eqref{eq:IndirectCond2}. 
In particular, the stationarity condition~$\mathcal{H}_u=\vec{0}$, viewed as an observable output in~\eqref{eq:observeDyn}, uniquely determines~$\vec{p}(t)$ for all~$t\in[0,T^\ast]$.
\end{lemma}

\begin{pf}

The constant costate~$q$ can be readily reconstructed through~\eqref{eq:qValue}:
\begin{equation}\label{eq:reconstruct_q}
    q = \frac{\partial c}{\partial y_T}\bigg\vert_{\big(T^\ast,\vec{x}^\ast(T^\ast),y(T^\ast),\bar{\vec{\sigma}}\big)},
\end{equation}
where $y(T^\ast)=\int_0^{T^\ast}l\big(\vec{x}^\ast(t),\vec{0},\bar{\vec{\sigma}}\big)\mr{d}t$.

Under Assumption~\ref{assumption:uniqueness}, the costates $\vec{p}(\cdot)$ are uniquely associated to $\vec{x}^\ast$ and $\vec{u}^\ast$ by~\eqref{eq:observeDyn} and are locally observable within~\eqref{eq:H_u} (cf. Section 1.9 in \cite{Isidori1995}). 
Exploiting the fact that $\vec{p}$ and $q$ are affine in the vector field~$\vec{\digamma}$ and in the function~$\mathcal{H}_u$, we use Assumption~\ref{assumption:smooth} and the Lie derivative~$\mathcal{L}$ to define
\begin{subequations}
\begin{equation}
    \left.\begin{bmatrix}\mathcal{H}_u\\\mathcal{L}_{\vec{\digamma}}\mathcal{H}_u\\\mathcal{L}_{\vec{\digamma}}\mathcal{L}_{\vec{\digamma}}\mathcal{H}_u\\\vdots
    \end{bmatrix}\right\vert_{\big(\vec{x}^\ast,\vec{p},\vec{u}^\ast,\bar{\vec{\sigma}}\big)}=\tilde{\vec{A}}\vec{p}+\tilde{\vec{b}}q\overset{!}{=}\vec{0},
\end{equation}
where
\begin{align}
    \tilde{\vec{A}}&=\begin{bmatrix}
        \frac{\partial \vec{f}}{\partial \vec{u}}\T\\
        \left(\sum_{i=1}^{n_\text{x}}f_i\frac{\partial }{\partial x_i} \frac{\partial \vec{f}}{\partial\vec{u}} \right)\T-\left(\frac{\partial \vec{f}}{\partial \vec{x}}\frac{\partial \vec{f}}{\partial \vec{u}}\right)\T\\ \vdots
    \end{bmatrix},\\
    \tilde{\vec{b}} &= \begin{bmatrix}
        \frac{\partial l}{\partial \vec{u}}\T\\
        \left(\sum_{i=1}^{n_\text{x}}f_i\frac{\partial }{\partial x_i} \frac{\partial l}{\partial\vec{u}} \right)\T-\left(\frac{\partial l}{\partial \vec{x}}\frac{\partial \vec{f}}{\partial \vec{u}}\right)\T\\ \vdots
    \end{bmatrix},\label{eq:bTilde}
\end{align}
\end{subequations}
are evaluated at the point~$(\vec{x}^\ast,\vec{u}^\ast,\bar{\vec{\sigma}})$.
Due to the local observability of $\vec{p}$, the matrix~ $\tilde{\vec{A}}$ has full (column) rank for all $\big(\vec{x}^\ast(t),\vec{u}^\ast(t)=\vec{0}\big)$ and $t\in[0,T^\ast]$.
Thus, for each time instant, we can construct a square matrix~$\vec{A}(t)\in\Real^{n_\mr{x}\times n_\mr{x}}$ from $n_\mr{x}$~rows of $\tilde{\vec{A}}$ such that $\text{det}(\vec{A}(t))\neq 0$.
Furthermore, selecting the same $n_\mr{x}$ rows from $\tilde{\vec{b}}$ constitutes the vector~$\vec{b}(t)\in\Real^{n_\mr{x}}$.
Finally, the costate can be reconstructed by
\begin{equation}\label{eq:reconstructCostate}
    \vec{p}(t) = \vec{A}(t)^{-1}\vec{b}(t)q,
\end{equation}
for any time~$t\in[0,T^\ast]$.
\end{pf}

\begin{proposition}[Reconstruction of Multipliers]\label{prop:multipliers}
Under the conditions of Lemma~\ref{lem:observe}, the Lagrange multipliers~$\vec{\lambda}$ associated with a passive-optimal solution are reconstructed from the transversality condition in \eqref{eq:IndirectCond2}.
\end{proposition}

\begin{pf}

By Lemma~\ref{lem:observe}, the boundary values of the costates, $\vec{p}(0)$ and $\vec{p}(T^\ast)$, can be reconstructed. Consequently, the remaining Lagrange multipliers~$\vec{\lambda}$ can be determined from~\eqref{eq:trans_x} as
\begin{align}
    \vec{\lambda}=&~\vec{H}^+ \cdot\bigg(\vec{p}(T^\ast)-\frac{\partial \vec{g}}{\partial \vec x_T}\bigg\vert_{\big(\vec{x}^\ast(T^\ast),\bar{\vec{\sigma}}\big)}\T\vec{p}(0)\bigg)\notag\\
       &~-\vec{H}^+ \cdot\bigg(\frac{\partial c}{\partial \vec x_T}\bigg\vert_{\big(T^\ast,\vec{x}^\ast(T^\ast),y(T^\ast),\bar{\vec{\sigma}}\big)}\T\bigg)\label{eq:reconstructLambda},\\
    \vec{H}=&~\bigg({\frac{\partial \vec{g}}{\partial \vec x_T}\hspace{-1mm}}\T {\frac{\partial \vec{h}}{\partial \vec{x}_0}\hspace{-1mm}}\T+{\frac{\partial \vec{h}}{\partial \vec x_T}\hspace{-1mm}}\T\bigg)\bigg\vert_{\big(T^\ast,\vec{x}^\ast(T^\ast),\vec{x}^\ast(0),\bar{\vec{\sigma}}\big)},
\end{align}
where $\vec{H}^+$ denotes the Moore–Penrose inverse of $\vec{H}$.

\end{pf}

\begin{remark}
    Full rank of $\vec{H}$, typical for passive-optimal
    legged motions~\citep{Raff2022b,Rosa2023}, ensures unique
    multiplier reconstruction via~\eqref{eq:reconstructLambda}
    and corresponds to the \emph{linear independence constraint
    qualification} for~\eqref{eq:boundCond1} and~\eqref{eq:indConst}.
    At isolated singularities, Lyapunov--Schmidt reduction
    can recover multipliers along branches extending from regular points.
\end{remark}

Constructing a parameterized optimization problem~$\mathcal{P}_{\vec{\sigma}}$ around a passive-optimal solution at $\bar{\vec{\sigma}}$ is particularly advantageous, as zero-input motions ($\vec{u} \equiv \vec{0}$) often globally minimize energy-based costs~$c$.
With reconstructed multipliers, these solutions provide
initial points for the continuation of optimal periodic motions.
\section{Implementation}
\label{sec:implementation}
We use single shooting to express the first-order optimality conditions~\eqref{eq:IndirectCond1}-\eqref{eq:IndirectCond2} of~$\mathcal{P}_{\vec{\sigma}}$ as a parameterized residual function. This formulation allows us to efficiently generate libraries of optimal gaits using numerical continuation.
To this end, we collect the unknown initial conditions and parameters into the decision variable
\begin{align}\label{eq:chiDef}
    \vec{\chi}\T =\begin{bmatrix}T & \vec{x}_0\T & \vec{p}_0\T & q & \vec{u}_0\T & \vec{\lambda}\T\end{bmatrix},    
\end{align}
where the initial input~$\vec{u}_0$ is an additional unknown
due to the semi-explicit form of the DAE~\eqref{eq:IndirectCond1}.
Let $\vec{\varphi}_\mr{w}:
\Real\times\Real^{n_\text{w}}\times\Real^{n_\sigma}
\to\Real^{n_\text{w}}$ denote the solution of~\eqref{eq:IndirectCond1} for $\vec{w}_0\T=[\vec{x}_0\T ~ 0 ~ \vec{p}_0\T ~ q ~ \vec{u}_0\T]$, where $n_\text{w}=2n_\text{x}+2+n_\text{u}$. We abbreviate $\vec{w}(t):=\vec{\varphi}_\mr{w}(t,\vec{w}_0;\vec{\sigma})$
and write the state and costate trajectories as
\begin{align}
    \vec{x}(t):=\vec{\varphi}_\mr{x}(t,\vec{w}_0;\vec{\sigma}),\quad
    \vec{p}(t):=\vec{\varphi}_\mr{p}(t,\vec{w}_0;\vec{\sigma}).
\end{align}

We compute $\vec{w}(t)$ over $t\in[0,T]$ using numerical
DAE solvers~\citep{Brenan1995,Hairer1989}.
Unlike ODE integration, solving DAEs requires additional care.
Even for smooth dynamics, solver choice and solution accuracy
depend strongly on how the algebraic condition
$\nicefrac{\partial \mathcal{H}}{\partial \vec{u}}=0$
is handled through index reduction.

\begin{assumption}[Trajectory Regularity]\label{assumption:regularParam}
For any initial condition $\vec{w}_0$ and parameter $\vec{\sigma}$, the trajectory exists on $[0,T]$ and is differentiable with respect to~$\vec{\sigma}$ and $\vec{\chi}$.
In particular, the functions $c$, $\vec{f}$, $l$, $\vec{g}$, and $\vec{h}$ are assumed to be continuously differentiable in $\vec{\sigma}$.
\end{assumption}

\begin{algorithm2e}[t]
\small
%\LinesNumbered
\DontPrintSemicolon %
\KwIn{Stationary point~$\vec{\chi}_\text{start}$ with parameter~$\sigma_\text{start}$ and desired range~$[\sigma_\text{start},\sigma_\text{end}]$; Step size $h>0$}
\KwOut{Gait family of stationary points~$\vec{\chi}$ of $\mathcal{P}_\sigma$ in the range~$[\sigma_\text{start},\sigma_\text{end}]$}
$\vec{\nu}_\text{start}=(\vec{\chi}_\text{start},\sigma_\text{start})$\;
compute initial tangent vector~$\vec{\tau}_\text{start}$ \Comment*[r]{eq. \eqref{eq:tangentvector}}
$d\in\{-1,1\}$ \Comment*[r]{eq.~\eqref{eq:Direction}}
$\vec{\nu}\gets\vec{\nu}_\text{start}$,~$\vec \tau\gets\vec \tau_\text{start}$,~$\sigma\gets\sigma_\text{start}$\; %
\vspace{1mm}
\While{$\sigma\neq \sigma_\text{end}$ } {
    \Comment*[l]{predictor}
    $\vec{\nu}_{\text{pred}} \gets \vec{\nu}+d h \vec \tau$ \Comment*[r]{eq. \eqref{eq:predictor}}
    compute $\vec \tau_\text{pred}$ at $\vec{\nu}_{\text{pred}}$ \Comment*[r]{eq. \eqref{eq:tangentvector}}
    \vspace{1.5mm}
    \Comment*[l]{corrector}
  \SetKwProg{Fn}{Newton's Method}{:}{}
  \Fn{$\mr{(}\vec{\nu}_\mr{pred},\vec{\tau}_\mr{pred}\mr{)}$}{
  \vspace{1mm}$\vec \nu \gets \vec \nu -\left[\begin{smallmatrix} {\mat{R}}(\vec \nu)\\ \vec{\tau}\T\end{smallmatrix}\right]^{-1}$\hspace{-0.5mm}$\cdot \left[\begin{smallmatrix}{\vec{r}}(\nu)\\[1mm]0\end{smallmatrix}\right]$\; 
        until convergence \Comment*[r]{loop}
        \KwRet $\left\{\vec \nu_\text{corr}=(\chi_\text{corr},\sigma_\text{corr});~\vec{\tau}_\text{corr}\right\}$
  }
$\vec{\nu}\gets\vec{\nu}_\text{corr}$,~$\vec \tau\gets \vec \tau_\text{corr}$,~$\sigma\gets\sigma_\text{corr}$\; 
}
\caption{Continuation to create 1D gait family}
\label{algo:OptimalContinuation}
\end{algorithm2e}

With this, we define the parameterized residual function~$\vec{r}:\Real^{N}\times \Real^{n_\sigma}\to\Real^{N}$, with $N=2n_\mr{x}+n_\mr{u}+4$, which collects the necessary conditions~\eqref{eq:IndirectCond1}-\eqref{eq:IndirectCond2} as:
\begin{subequations}\label{eq:r_ParamIndirect}
\begin{equation}
    \vec{r}(\vec{\chi},\vec{\sigma})=\begin{bmatrix}
        \vec{r}_{T}(\vec{\chi};\vec{\sigma})\\
       \vec{r}_{\vec{x}_T}(\vec{\chi};\vec{\sigma})\\
        q-\frac{\partial c}{\partial y_T}\big\vert_{\big(T,\vec{x}(T),y(T);\vec{\sigma}\big)}\\
        \frac{\partial \mathcal{H}}{\partial \vec{u}}\big\vert_{\big(\vec{w}(0);\vec{\sigma}\big)}\\
       \vec{g}(\vec{x}(T);\vec{\sigma})-\vec{x}_0\\
        \vec{h}\big(T,\vec{x}(T),\vec{x}_0;\vec{\sigma}\big)
    \end{bmatrix}=\vec{0}, 
\end{equation}
where
\begin{align}
    r_{T} =&~
       \mathcal{H}(\vec{w}(T);\vec{\sigma})+\frac{\partial c}{\partial T}\big\vert_{\big(T,\vec{x}(T),y(T);\vec{\sigma}\big)}\notag\\
       &~+\vec{\lambda}\T \frac{\partial \vec{h}}{\partial T}\big\vert_{\big(T,\vec{x}(T),\vec{x}_0;\vec{\sigma}\big)},\\
    \vec{r}_{\vec{x}_T}=&~\frac{\partial \vec{g}}{\partial \vec x_T}\bigg\vert_{\big(\vec{x}(T);\vec{\sigma}\big)}\T\cdot\left(\vec{p}_0+\frac{\partial \vec{h}}{\partial \vec{x}_0}\bigg\vert_{\big(T,\vec{x}(T),\vec{x}_0;\vec{\sigma}\big)}\T\vec{\lambda}\right)\notag\\
       &~-\vec{p}(T)+ \frac{\partial c}{\partial \vec x_T}\bigg\vert_{\big(T,\vec{x}(T),y(T);\vec{\sigma}\big)}\T\notag\\
       &~+\frac{\partial \vec{h}}{\partial \vec x_T}\bigg\vert_{\big(T,\vec{x}(T),y(T);\vec{\sigma}\big)}\T\vec{\lambda}.
\end{align}
\end{subequations}
For a fixed parameter vector~$\bar{\vec{\sigma}}$, we define the residual~$\bar{\vec{r}}:\Real^{N}\to\Real^{N}$ such that $\bar{\vec{r}}(\vec{\chi}):=\vec{r}(\vec{\chi},\bar{\vec{\sigma}})$.
Imposing Assumption~\ref{assumption:regularParam}, we further define the Jacobians:
\begin{subequations}
    \begin{align}
        \bar{\vec{R}}(\vec{\chi})&:=\frac{\partial \bar{\vec{r}}}{\partial \vec{\chi}}\in\Real^{N\times N},\\
        \vec{R}(\vec{\chi},\vec{\sigma})&:=\begin{bmatrix}
            \frac{\partial \vec{r}}{\partial \vec{\chi}} & \frac{\partial \vec{r}}{\partial {\vec{\sigma}}}
        \end{bmatrix}\in\Real^{N\times (N+n_\sigma)}.
    \end{align}
\end{subequations}
The computation of these Jacobians is crucial for Newton-type solvers employed in numerical continuation methods.
In this paper, we restrict ourselves to one-dimensional gait families ($n_\sigma=1$), such that the residual function~\eqref{eq:r_ParamIndirect} implicitly defines a curve of stationary solutions.
To trace this curve, predictor-corrector methods are particularly well-suited due to their strong local convergence properties.
We therefore employ a pseudo-arclength continuation method adapted from~\cite{Allgower2003}, which iteratively predicts a new point along the solution curve and corrects it using Newton's method.
Starting from a known stationary solution, this procedure
generates families of optimal gaits over a prescribed parameter range, as summarized in Algorithm~\ref{algo:OptimalContinuation}.
Further details on the predictor-corrector method are provided in Appendix~\ref{sec:appendixContinuation}.

\begin{remark}   
    Algorithm~\ref{algo:OptimalContinuation} assumes monotonicity of the solution curve~${\vec{r}}^{-1}(\vec{0})$ in the parameter~$\sigma\in[\sigma_\text{start}, \sigma_\text{end}]$ to determine the continuation direction~$d$. 
    However, along the solution curve, turning points in~$\sigma$ (Section~\ref{sec:levelground}) or bifurcation points, where the gait family branches into multiple solution paths, may be encountered. Consequently, reaching~$\sigma_\text{end}$ from an initial stationary point~$\vec{\chi}_\text{start}$ is not guaranteed.
\end{remark}

\section{The Compass-Gait Walker}
\label{sec:example}
To demonstrate the effectiveness of the proposed methods in legged robotics, we computed optimal gaits for an underactuated compass-gait walker (CGW) and compared the results with established methods based on direct trajectory optimization~\cite{Wensing2024,Diehl2006,Rosa2023}. Our analysis focuses on both computational efficiency and the numerical sensitivity of the different formulations. 
Since the indirect method does not rely on a finite parameterization of the control inputs, it serves as a reliable benchmark for assessing various input parameterizations used in direct methods.
Despite the simplicity of the CGW which has only four state variables ($n_\mr{x}=4$), it remains a widely adopted model for studying bipedal locomotion~\cite{Goswami1997, Spong1999, Asano2004, Manchester2011, Westervelt2007, Rosa2014, Rosa2023}.
Its characteristics—underactuation, non-smooth dynamics, limit cycle behavior, and the need for optimizing energy efficiency—are representative of the challenges encountered in more complex legged robots.
Additionally, the compass-gait walker (Figure~\ref{fig:cgw}) can passively walk down a slope driven solely by gravity~$g$. We used this capability to generate gait families parameterized by the slope's incline and the robot's average speed.

\begin{figure}[t]
    \centering
    \resizebox{\linewidth}{!}{\includegraphics{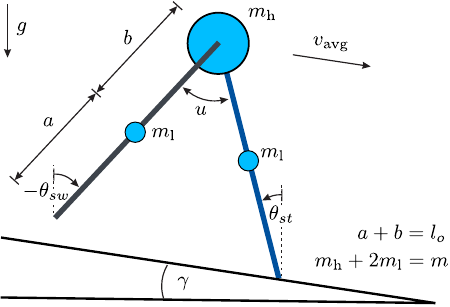}}
    \caption{The compass-gait walker on a slope~$\gamma$, with parameters normalized by total mass~$m$, gravity~$g$, and leg length~$l_\mr{o}$, and proportions $\nicefrac{a}{b}=1$ and $\nicefrac{m_\mr{h}}{m_\mr{l}}=2$.}
    \label{fig:cgw}
\end{figure}
\begin{figure*}[t]
    \centering
    \begin{subfigure}[t]{\columnwidth}
        \centering
        \includegraphics{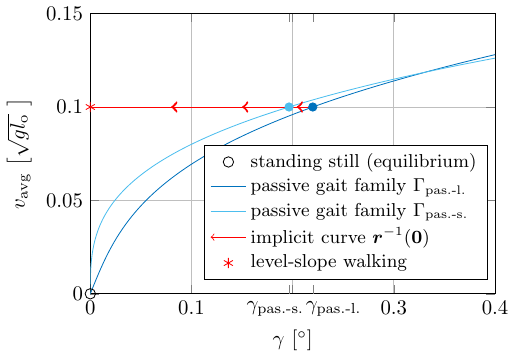}
        \caption{Continuation process towards level-ground walking, crossing two passive gaits.}
        \label{fig:subpassive}
    \end{subfigure}\hfill
    \begin{subfigure}[t]{\columnwidth}
        \centering
        \includegraphics{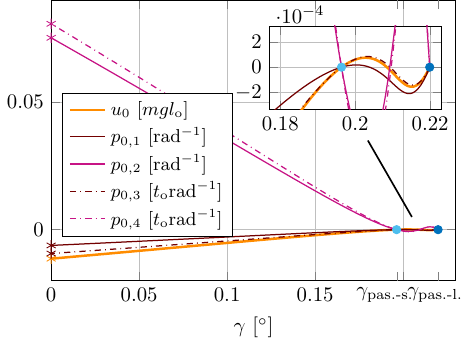}
        \caption{Evolution of the initial input~$u_0$ and costates~$\vec{p}_0$ within $\vec{r}^{-1}(\vec{0})$. At passive gaits ($u\equiv 0$), the costates vanish ($\vec{p}\equiv \vec{0}$).}
        \label{fig:subgamma}
    \end{subfigure}
    \caption{Projections of the implicit curve~$\vec{r}^{-1}(\vec{0})$ for the compass-gait walker (CGW), highlighted in varying shades of red and computed using Algorithm~\ref{algo:OptimalContinuation} with continuation parameter~$\sigma := \gamma$ and fixed average speed~$v_\text{avg} = 0.1~\sqrt{gl_\text{o}}$. Towards level ground~$\gamma = 0^\circ$ (marked by $\ast$), the curve~$\vec{r}^{-1}(\vec{0})$ connects the passive gait families with longer ($\Gamma_\text{pas.-l.}$) and shorter ($\Gamma_\text{pas.-s.}$) periods at slopes~$\gamma_\text{pas.-l.}$ and~$\gamma_\text{pas.-s.}$, respectively. Filled circles mark where passive and optimal gaits coincide.}
    \label{fig:passive2active}
\end{figure*}

The model consists of a stance leg (with angle $\theta_\text{st}$ relative to vertical) that remains on the ground without sliding and a swing leg~(with angle $\theta_\text{sw}$ relative to vertical) that is allowed to swing freely.
During forward motion, the swing leg strikes the ground ahead of the stance leg, followed by an instantaneous switch between the swing and stance legs.
Hence, the swing leg becomes the stance leg and vice versa.
Both legs have a length~$a+b=l_\mr{o}$ with a point mass~$m_\mr{l}$ located at a distance $a$ below the common hip joint.
The point mass~$m_\mr{h}$ at the hip constitutes the main body of the biped.
A single motor~($n_\mr{u}=1$) is located at the hip to provide input torques~$u$ that act between the two legs.
The robot's specific proportions are provided in Figure~\ref{fig:cgw}, where we normalized its parameters by total mass~$m$, gravity~$g$, and leg length~$l_\mr{o}$.
Specifically, normalized time is defined as $t_\mr{o}=\sqrt{l_\mr{o}/g}$.
Its state is~$\vec{x}=[\vec{q}~\dot{\vec{q}}]\T$ with configuration~$\vec{q}=[\theta_\mr{sw}~\theta_\mr{st}]\T$.
The continuous dynamics~$\vec{f}(\vec{x},u)$ and discrete dynamics~$\vec{g}(\vec{x})$ are reported in Appendix~\ref{sec:cgw}.
Due to this choice of minimal coordinates, the dynamics are independent of the slope~$\gamma$.
However, the slope affects the final event of both legs touching the ground:
\begin{subequations}\label{eq:eqConCGW}
\begin{equation}\label{eq:eventCGW}
    e=\theta_\mr{sw}(T)+\theta_\mr{st}(T)+2\gamma.
\end{equation}
In addition to periodicity, we required the robot to walk with a given average speed~$v_\text{avg}$, which is specified within a single operating condition ($n_\omega=1$):
\begin{equation}\label{eq:operatingCGW}
    \omega=2\sin\big(\theta_\mr{sw}(T)+\gamma\big)-v_\text{avg}T.
\end{equation}
\end{subequations}
The equality constraint~$\vec{h}\big(T,\vec{x}(T),\vec{x}(0);\vec{\sigma}\big)=\vec{0}$ of the parameterized problem~$\mathcal{P}_{\vec{\sigma}}$ follows from \eqref{eq:eventCGW} and~\eqref{eq:operatingCGW}, where $\vec{\sigma}\T=[\gamma~v_\text{avg}]$.
The objective in this example was to minimize the cost of transport, defined as
\begin{equation}
    c = \frac{y(T)}{mg\Delta x}\quad\Bigg\vert 
     \quad\dot{y} =\underbrace{\tfrac{k}{2}~u^2}_{=l(\vec{x},u)},~\Delta x = v_\text{avg}T,
\end{equation}
where~$k=(m\sqrt{gl_\mr{o}^3})^{-1}$ ensures a dimensionless cost of transport~$c$ without affecting the optimum of~$\mathcal{P}$.
The traveled distance~$\Delta x$ is typically constrained to be strictly positive and bounded, ensuring meaningful forward gaits.
Consequently, the cost attains its minimum of zero only for passive-optimal gaits with~$u\equiv 0$.
\begin{remark}\label{rem:uSquared}
With the instantaneous cost defined as $l=\nicefrac{k}{2}~u^2$,
the reconstruction of Lagrange multipliers at passive optimal gaits (Section~\ref{sec:reconstructLagrange}) becomes straightforward, as the vector~$\tilde{\vec{b}}$ from equation~\eqref{eq:bTilde} vanishes. Consequently, from equation~\eqref{eq:reconstructCostate}, the costate~$\vec{p}$ is zero at all times ($\vec{p} \equiv \vec{0}$). 
By inspection of equation~\eqref{eq:reconstructLambda}, the Lagrange multiplier~$\vec{\lambda}$ is also zero ($\vec{\lambda} = 0$). 
The remaining scalar costate~$q$ is reconstructed using equation~\eqref{eq:reconstruct_q}, yielding 
$q = (mgv_\text{avg} T^\ast)^{-1}$.
\end{remark}

In the following, we leveraged the passive optimal gaits of the CGW to initialize Algorithm~\ref{algo:OptimalContinuation}, generating optimal families of gaits within the parameterized optimization problem~$\mathcal{P}_{\vec{\sigma}}$ (Section~\ref{sec:passive2Level}). 
Additionally, this workflow enabled a performance and accuracy comparison between the indirect and direct methods (Section~\ref{sec:indVSdir}).
This comparison was also extended to gait families that do not include passive gaits (Section~\ref{sec:levelground}).

Algorithm~\ref{algo:OptimalContinuation} was implemented in MATLAB with a residual tolerance of $10^{-8}$. We used MATLAB's variable step-size integrator \verb"ode15s", with relative and absolute tolerances of $10^{-9}$ and $10^{-10}$, respectively, to solve the DAEs within ${\vec{r}}$. 
To ensure a fair comparison, the same settings were applied for solving the ODEs in the direct method. 
The Jacobians~$\vec{R}$ were approximated by forward finite differences with a step size of $10^{-9}$.
The corresponding code is available on \href{https://github.com/raffmax/IndirectMethod}{GitHub}\footnote{\href{https://github.com/raffmax/IndirectMethod}{\url{github.com/raffmax/IndirectMethod}}}.

\subsection{From Passive to Level Ground Walking}\label{sec:passive2Level}
To demonstrate the exploratory potential of the indirect approach combined with numerical continuation, we generated a library of walking gaits on a range of slopes.
We used the scalar continuation parameter~$\sigma := \gamma$ while fixing the average speed at~$v_\text{avg} = 0.1 \sqrt{gl_\text{o}}$. Algorithm~\ref{algo:OptimalContinuation} was initialized with the passive CGW gait at slope~$\gamma_\text{pas.-l.}$ (dark blue dot in Figure~\ref{fig:passive2active}a), identified through a continuation process similar to that in \cite{Rosa2023} (see Section~\ref{sec:passiveConstruction}). This gait served as the optimal passive solution~$(T_\text{pas.-l.}^\ast, \vec{x}_\text{pas.-l.}^\ast(\cdot), u_\text{pas.-l.}^\ast(\cdot))$, with zero input~$u_\text{pas.-l.}^\ast \equiv 0$ and cost~$c = 0$ for the parameterized optimization problem~$\mathcal{P}_{\gamma}$.

As noted in Remark~\ref{rem:uSquared}, the reconstruction of the costates and Lagrange multipliers is straightforward. Thus, Algorithm~\ref{algo:OptimalContinuation} was initialized with
\begin{subequations}
\begin{align}
    \vec{\chi}_\text{start} &= \begin{bmatrix}
        T_\text{start} \\
        \vec{x}_{0,\text{start}} \\
        \vec{p}_{0,\text{start}} \\
        q_\text{start} \\
        u_{0,\text{start}} \\
        \vec{\lambda}_\text{start}
    \end{bmatrix} = \begin{bmatrix}
        T_\text{pas.-l.}^\ast \\
        \vec{x}_\text{pas.-l.}^\ast(0) \\
        \vec{0} \\
        (mgv_\text{avg} T_\text{pas.-l.}^\ast)^{-1} \\
        0 \\
        \vec{0}
    \end{bmatrix}, \\
    \sigma_\text{start} &= \gamma_\text{pas.-l.},
\end{align}
\end{subequations}
where $v_\text{avg} = 0.1 \sqrt{gl_\text{o}}$ was fixed throughout the algorithm.
To generate a gait family that includes level-ground walking, the desired range of the slope parameter was set to $[\sigma_\text{start}, \sigma_\text{end}] = [\gamma_\text{pas.-l.}, 0^\circ]$. 
Figure~\ref{fig:passive2active}a shows a projection of the gait family curve $\vec{r}^{-1}(\vec{0})$, depicted in red, as the output of Algorithm~\ref{algo:OptimalContinuation}. 
The red curve also indicates an intersection with another passive gait at slope~$\gamma_\text{pas.-s.}$, characterized by a shorter period (light blue dot in Figure~\ref{fig:passive2active}a). 
In Figure~\ref{fig:passive2active}b, we further illustrate how the initial costates~$\vec{p}_0$ and input~$u_0$ vanish at those slopes where passive gait exists.

\begin{figure*}[t]
    \centering
    \begin{minipage}{\columnwidth}
        \centering
    \begin{subfigure}{\columnwidth}
        \centering
        \includegraphics[width=0.98\textwidth]{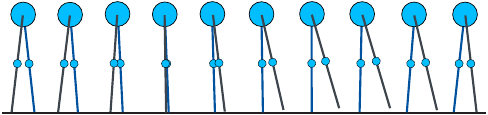}
    \end{subfigure}\\[4mm]
    \begin{subfigure}{\columnwidth}
        \centering
        \includegraphics{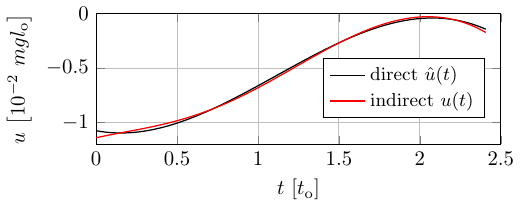}
    \end{subfigure}
    \end{minipage}\hfill
    \begin{minipage}{0.95\columnwidth}
    \begin{subfigure}{\columnwidth}
        \centering
        \includegraphics{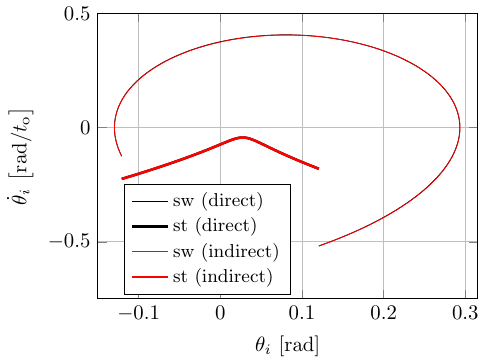}
    \end{subfigure}
    \end{minipage}
    \caption{Visualization of an optimal walking gait of the CGW at an average speed of $v_\text{avg} = 0.1 \sqrt{gl_\text{o}}$ on level ground ($\gamma=0^\circ$). Keyframes illustrate the walking motion, while input and state trajectories (swing leg "sw", stance leg "st") compare indirect~(red) and direct (black) shooting methods. The direct method parameterizes the input with a single cubic B-spline ($n_\xi=4$).}
    \label{fig:visualGait}
\end{figure*}

\subsubsection{Construction of Passive Gait Libraries}\label{sec:passiveConstruction}
The two passive gaits at $\gamma_\text{pas.-s.}$ and $\gamma_\text{pas.-l.}$ do not have to be found by trial and error.  Instead, they can be systematically obtained from the standing equilibrium via pseudo-arclength continuation \citep{Rosa2014}, using a procedure analogous to Algorithm~\ref{algo:OptimalContinuation}.
The key to this systematic approach is to consider an equilibrium as a periodic motion with a continuous range of period times. For each period, it can then be determined whether a local change in $\gamma$ would result in a passive downhill walk. This is accomplished through the systematic detection of a so-called simple bifurcation, followed by a Lyapunov–-Schmidt reduction to leave the degenerate periodic motion~\citep{Allgower2003}.
For the CGW, the first two simple bifurcation points along increasing period at standing still generate two passive gait families with shorter ($\Gamma_\text{pas.-s.}$) and longer ($\Gamma_\text{pas.-l.}$) periods.
The result of this continuation process is shown by the blue lines in Figure~\ref{fig:passive2active}a.

\subsection{Comparison to the Direct Method}\label{sec:indVSdir}
Building on the gait family~$\vec{r}^{-1}(\vec{0})$ computed in the previous section, we performed a comparison between the indirect method and the direct method. 
The direct method, combined with a variable step-size shooting approach (MATLAB's~\verb"ode15s"), requires a parameterization~$\vec{\xi} \in \Real^{n_\xi}$, with~$n_\xi>1$, of the input space, where the control input is defined as~$\vec{u} := \hat{\vec{u}}(t, \vec{\xi})$.
This readily allows to numerically solve the differential equation~\eqref{eq:indDynOpt1} and yields the flow~$\hat{\vec{x}}(t):=\vec{\varphi}_{\mr{x}}(t,\vec{x}_0,\vec{\xi})$.
We implemented and compared cubic B-splines and Bézier curves
for varying $n_\xi$, both offering high accuracy with few
parameters~\citep{Lin2014}.
A detailed description of the input parameterizations and our implementation of the direct method is provided in Appendix~\ref{sec:appendixDirect}.
By transcribing the input space of problem~$\mathcal{P}_{\vec{\sigma}}$ with $\vec{u}:=\hat{\vec{u}}(t,\vec{\xi})$, we pose the direct shooting problem as an implicit function, similar to the parameterized indirect method in~\eqref{eq:r_ParamIndirect}.
Hence, the residual~$\hat{\vec{r}}:\Real^{\hat{N}}\times\Real^{n_\sigma}\to\Real^{\hat{N}}$ of the direct method, with $\hat{N}=2n_\mr{x}+n_\mr{\xi}+3$, is defined as
\begin{subequations}\label{eq:rHatgamma}
\begin{align}
    \hat{\vec{r}}(\hat{\vec{\chi}},\vec{\sigma})&=\begin{bmatrix}
        \frac{\partial \hat{c}}{\partial \vec{s}}\big\vert_{(\hat{\vec{s}},{\vec{\sigma}})}\T+\frac{\partial \hat{\vec{h}}}{\partial \vec{s}}\big\vert_{(\hat{\vec{s}},{\vec{\sigma}})}\T\hat{\vec{\lambda}}\\
        \hat{\vec{h}}\vert_{(\hat{\vec{s}},{\vec{\sigma}})}
    \end{bmatrix}=\vec{0},\label{eq:rHat}\\
    \hat{\vec{\chi}}\T &= \bigg[\underbrace{\big[T \quad \vec{x}_0\T \quad \vec{\xi}\T\big]}_{=:\hat{\vec{s}}\T}\quad \hat{\vec{\lambda}}\T
    \bigg],
\end{align}
where
\begin{align}
    \hat{c}(\hat{\vec{s}},{\vec{\sigma}}) &= c(T, \vec{x}(T), y(T);\vec{\sigma}),\\
    \hat{\vec{h}}(\hat{\vec{s}},{\vec{\sigma}}) &=\begin{bmatrix}
        \vec{g}(\hat{\vec{x}}(T))-\vec{x}_0\\
        \vec{h}\big(T,\hat{\vec{x}}(T),\vec{x}_0;{\vec{\sigma}}\big)
        \end{bmatrix} ,\label{eq:hHat}
\end{align}
\end{subequations}
and $\hat{\vec{\lambda}}\in\Real^{n_\mr{x}+2}$ is the associated Lagrange multiplier to the constraint~\eqref{eq:hHat}.
Similar to the indirect case, when the parameters~$\bar{{\vec{\sigma}}}$ are fixed, we refer to the residual~\eqref{eq:rHatgamma} as $\hat{\bar{\vec{r}}}\in\Real^{\hat{N}}\to \Real^{\hat{N}}$.
And likewise, its respective Jacobians are denoted as $\hat{\vec{R}}\in\Real^{\hat{N}\times\hat{N}+n_\sigma}$ and $\hat{\bar{\vec{R}}}\in\Real^{\hat{N}\times\hat{N}}$.

In accordance with \cite{Rosa2023}, we utilized the direct shooting implementation of the residual~\ref{eq:rHatgamma} and Algorithm~\ref{algo:OptimalContinuation} to generate a family of gaits~$\hat{\vec{r}}^{-1}(\vec{0})$, which includes both passive gaits and a level-ground walking gait in the parameter range~$[\sigma_\text{start}, \sigma_\text{end}] = [\gamma_\text{l.}, 0^\circ]$.
As in the indirect shooting method described in Section~\ref{sec:passive2Level}, we defined the scalar continuation parameter as $\sigma:=\gamma$, while constraining the average speed to $v_\text{avg}=0.1\sqrt{gl_\text{o}}$.

\begin{figure*}[t]
	\centering
	\begin{subfigure}[b]{1\columnwidth}
		\centering
		\includegraphics{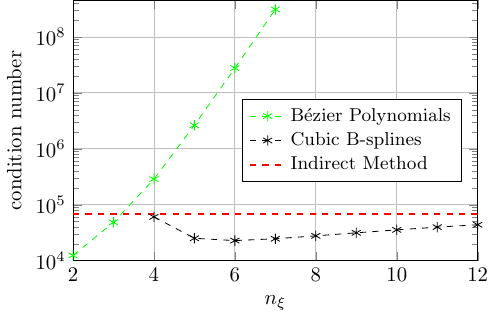}
		\caption{Condition numbers of the indirect Jacobian ($\bar{\vec{R}}$) and direct Jacobian ($\hat{\bar{\vec{R}}}$). The direct method using Bézier polynomials fails to converge for~$n_\xi > 7$ due to high sensitivity in~$\vec{\xi}$ within the shooting problem.}
	\end{subfigure}\hfill
	\begin{subfigure}[b]{1\columnwidth}
		\centering
		\includegraphics{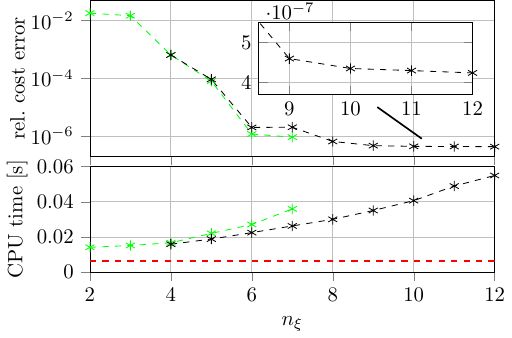}
		\caption{Relative error $\frac{\hat{c} - c}{c}$ in the optimal cost between the direct method ($\hat{c}$) and the indirect method ($c$), along with the average CPU time over 1000 evaluations of the indirect function $\bar{\vec{r}}$ and the direct function $\hat{\bar{\vec{r}}}$.}
		\label{fig:CPU}
	\end{subfigure}
    \caption{Comparison of direct ($\hat{\bar{\vec{r}}}$) and indirect ($\bar{\vec{r}}$) approaches for the CGW, implemented in MATLAB on an Intel Core i5-8500 (3.00 GHz, 8 GB RAM). The direct method parameterizes the input using Bézier polynomials or cubic B-splines with $n_\xi$ variables. Results in (a) and (b) correspond to level-ground walking ($\gamma=0^\circ$) at $v_\text{avg}=0.1\sqrt{gl_\text{o}}$. The corresponding (locally) optimal gaits were generated via continuation from a passive gait using Algorithm~\ref{algo:OptimalContinuation}, as shown in Figure~\ref{fig:passive2active}.}
	\label{fig:directVSindirect}
\end{figure*}
Figures~\ref{fig:visualGait} and \ref{fig:directVSindirect} compare the direct~($\hat{\bar{\vec{r}}}$) and indirect~($\bar{\vec{r}}$) shooting methods for a locally optimal gait at level slope~($\gamma=0^\circ$) and average speed~$v_\text{avg} = 0.1 \sqrt{gl_\text{o}}$. 

To analyze and contrast the region of convergence, we computed the condition numbers of $\bar{\vec{R}}$ and $\hat{\bar{\vec{R}}}$ for varying numbers of input parameters~$n_\xi$ (Figure~\ref{fig:directVSindirect}a). 
While the condition number of the indirect method remains constant, as it is independent of the input parameterization, the condition number of $\hat{\bar{\vec{R}}}$ varies with~$n_\xi$. 
The control input parameterization using Bézier polynomials becomes highly sensitive to parameter variations as $n_\xi$ increases. 
This sensitivity arises from the global nature of this parameterization, where adjusting a single parameter~$\xi_i$ in the vector~$\vec{\xi}$ affects the entire shape of the input curve~$\hat{\vec{u}}(t,\vec{\xi})$. 
As a result, for $n_\xi > 7$, Algorithm~\ref{algo:OptimalContinuation} stalled at certain points in traversing $\hat{\vec{r}}^{-1}(\vec{0})$, where Newton's method failed to converge due to the high sensitivities and the use of forward finite differences in computing~$\hat{\vec{R}}$.

In contrast, the cubic B-spline parameterization introduces $n_\xi-3$ equidistant segments in the input space, which are locally decoupled. 
Thus, a variation in a single parameter~$\xi_i$ only locally affects the input curve~$\hat{\vec{u}}(t,\vec{\xi})$. 
This allows for arbitrarily large parameterizations without a blow-up in the condition number, as illustrated in Figure.~\ref{fig:directVSindirect}a. 
Therefore, as the accuracy of approximating the control input space increases, cubic B-splines are much better suited. 
This is also evident in Figure~\ref{fig:directVSindirect}b, which shows the relative cost error of direct shooting~($\nicefrac{\hat{c}-c}{c}$) compared to the indirect cost~($c$). 
Note that the cost value $\hat{c}$ of the cubic B-spline parameterization approaches the indirect cost value $c$ as $n_\xi$ increases, although it remains inferior. 
It is also worth noting that Bézier polynomials and cubic B-splines yield the same cost value at $n_\xi = 4$ since both describe a cubic curve. 
However, the CPU time required to evaluate~$\hat{\vec{r}}$ for these parameterizations differs slightly, as cubic B-splines are more efficiently computed in a recursive manner (Appendix~\ref{sec:inputTransc}). 
This plot of CPU times also highlights that indirect shooting can be significantly faster than direct shooting when $n_\xi$ becomes large.

There is, of course, considerable room for improving efficiency, as the current implementation is in MATLAB. 
In addition to optimizing runtime by transitioning to a more efficient programming language such as Julia or C++, there is potential for accelerating both the direct and indirect shooting approaches. 
Specifically, there are alternative approaches for computing the first-order condition within \eqref{eq:rHat} in the direct method, as detailed in Appendix~\ref{sec:sens}. 
Likewise, the indirect shooting approach can be significantly accelerated by eliminating the input variable~$u_0$ from $\vec{\chi}$. 
Often, the control input~$u$ can be explicitly expressed as a function of $\vec{x}$, $\vec{p}$, and $q$ by utilizing $\nicefrac{\partial\mathcal{H}}{\partial u}=0$. 
With an instantaneous cost~$l=\nicefrac{k}{2}u^2$ and the input-affine dynamics of the CGW, the input takes the form:
\begin{equation}\label{eq:CGWu}
    u = -\frac{1}{kq}\frac{\partial \vec{f}}{\partial u}\bigg\vert_{\vec{x}(t)}\T \cdot \vec{p}(t).
\end{equation}
By eliminating the variable~$u_0$, this step transforms the DAE into an ODE in $\vec{x}$ and $\vec{p}$, which can generally be solved more quickly. In Section~\ref{sec:RABBIT}, this approach accelerates dynamics and gradient computations for the higher-dimensional RABBIT system.
The transformation also enables the use of variable step-size symplectic integrators~\citep{Hairer1997}, which effectively exploit the Hamiltonian structure.
Such integrators enhance conservation properties throughout the simulation and can be particularly beneficial in addressing shooting problems when the period~$T$ is large.
However, these refinements, including a detailed examination of the computational efficiency trade-offs between direct and indirect shooting methods, are beyond the scope of this work and will be addressed in future research.

\subsection{Generation of Additional Gait Families}\label{sec:levelground}
\begin{figure}[t]
    \centering
    \includegraphics{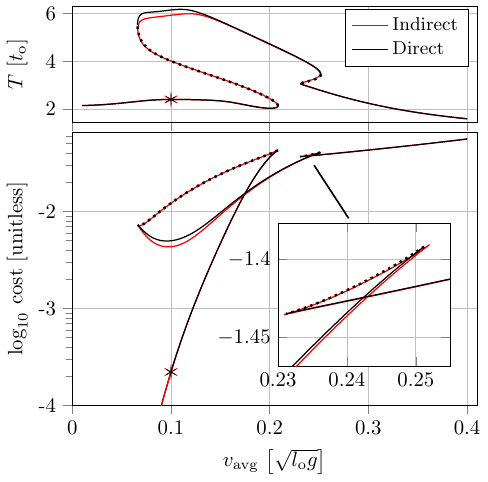}
    \caption{Projections of the implicit curves from the indirect ($\vec{r}^{-1}(\vec{0})$) and direct ($\hat{\vec{r}}^{-1}(\vec{0})$) methods, computed using Algorithm~\ref{algo:OptimalContinuation} with continuation parameter~$\sigma = v_\text{avg}$. Both curves describe level-ground walking ($\gamma=0^\circ$), starting from an optimal gait at $v_\text{avg}=0.1~\sqrt{gl_\text{o}}$ (marked by $\ast$). The indirect method consistently achieves lower costs, while the direct method (utilizing cubic B-splines with $n_\xi=4$) distinguishes local minima (solid black) from saddle points (dotted black) using second-order optimality conditions.}
    \label{fig:levelground}
\end{figure}
In addition to the previous sections, we further demonstrated the versatility and practical applicability of the proposed methods in legged locomotion by generating an additional family of optimal gaits that do not directly stem from passive dynamics.
Specifically, we used the previously computed locally optimal gait at level slope~($\gamma=0^\circ$) and average speed~$v_\text{avg}=0.1\sqrt{gl_\text{o}}$ to initialize Algorithm~\ref{algo:OptimalContinuation} for the generation of a family of level walking gaits~$\vec{r}_{v_\text{avg}}^{-1}(\vec{0})$.
Unlike earlier sections where we varied the slope, we fixed the slope at~$\gamma=0^\circ$ and varied the speed, using it as our continuation parameter~$\sigma=v_\text{avg}$. This approach underscored the method's ability to adapt to different operating conditions, extending its relevance to practical robotic systems.

As shown in Figure~\ref{fig:levelground}, the gait family (red curve) was generated by running Algorithm~\ref{algo:OptimalContinuation} twice, with $\sigma_\text{end}=0.01\sqrt{gl_\text{o}}$ and $\sigma_\text{end}=0.4\sqrt{gl_\text{o}}$.
In the latter case, tracing $\vec{r}_{v_\text{avg}}^{-1}(\vec{0})$ encounters four turning points in $v_\text{avg}$, yielding multiple stationary gaits at certain speeds.
Without second-order conditions, the indirect method cannot distinguish local minima from saddle points.
Deriving such conditions is cumbersome even for restricted classes of periodic optimization problems~\citep{Speyer1984}, requiring, among other things, particular solutions to Riccati differential equations and analysis of the monodromy matrix of the underlying DAE/ODE.

While deriving second-order conditions for $\mathcal{P}$ is out of scope for this work, we leveraged the direct method for comparison, as these conditions are more accessible (see Appendix~\ref{sec:2ndCond}).
The black curve in Figure~\ref{fig:levelground} shows projections from the direct shooting approach (points in $\hat{\vec{r}}_{v_\text{avg}}^{-1}$) using a single cubic B-spline~($n_\xi=4$).
Solid lines indicate strict local minimizers, while dashed lines correspond to families of saddle points.
When traversing the curve~$\hat{\vec{r}}_{v_\text{avg}}^{-1}$, the switch between minimizer and saddle point always occurs at turning points in $v_\text{avg}$.

Furthermore, examining the cost plot in Figure~\ref{fig:levelground}, we could distinguish between good and bad local minima.
Notably, similar to Figure~\ref{fig:directVSindirect}b, the indirect method consistently yielded a lower cost value.
This is due to the input space parameterization in the direct method, which can only approximate the true optimal solution, whereas the indirect method is more flexible, utilizing an adaptive step-size DAE solver.

\section{RABBIT}
\label{sec:RABBIT}
\begin{figure}[t]
	\centering
\includegraphics[width=\linewidth]{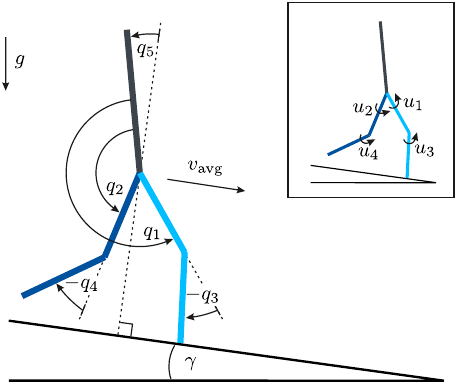}
    \caption{The RABBIT robot on a slope~$\gamma$, with parameters taken from~\citep{Chevallereau2003,Westervelt2007} normalized by total mass~$m$, gravity~$g$, and leg length~$l_\mr{o}$.}
	\label{fig:rabbit}
\end{figure}
\begin{figure*}[t]
	\centering
	\begin{subfigure}{\columnwidth}
		\centering
		\includegraphics{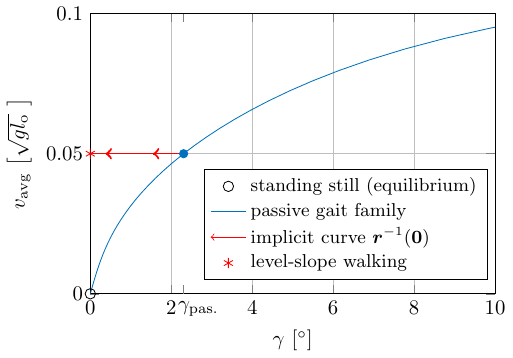}
		\caption{Continuation process towards level-ground walking, originating from a passive gait.}
		\label{fig:subpassiveRABBIT}
	\end{subfigure}\hfill
	\begin{subfigure}{\columnwidth}
		\centering
        \includegraphics{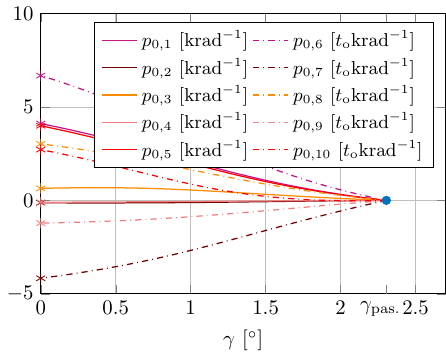}
	    \caption{Evolution of the initial costates~$\vec{p}_0$ ($\mr{krad}=10^{3}\mr{rad}$) along~$\vec{r}^{-1}(\vec{0})$, vanishing at the passive gait ($\vec{u}\equiv\vec{0}$).}
		\label{fig:subgammaRABBIT}
	\end{subfigure}
	\caption{Projections of the implicit curve $\vec{r}^{-1}(\vec{0})$ for the robot RABBIT, computed using Algorithm~\ref{algo:OptimalContinuation} with the continuation parameter~$\sigma := \gamma$. 
		The curve corresponds to a fixed average speed of $v_\text{avg} = 0.05~\sqrt{gl_\text{o}}$ and originates from a passive gait family at slope~$\gamma_\text{pas}$. 
		The filled circle indicates the point where the passive and optimal gait coincide.}
	\label{fig:passive2activeRABBIT}
\end{figure*}

To demonstrate that the methodology for generating optimal gait families (Algorithm~\ref{algo:OptimalContinuation}) extends to more complex systems, we applied the indirect method to a model of the legged robot RABBIT. As depicted in Figure~\ref{fig:rabbit}, the robot has ten state variables ($n_\text{x}=10$) in minimal coordinates, with one foot always in contact with the ground. It is actuated by four motors ($n_\text{u}=4$) and is therefore underactuated.
Nearly all physical parameters of RABBIT are taken from \citep{Chevallereau2003,Westervelt2007} and normalized by the total mass $m$, gravity~$g$, and leg length~$l_\mr{o}$.
The only exception is joint friction, which was removed in order to allow RABBIT to passively walk down a shallow slope driven solely by gravity $g$, similar to the CGW. 
As noted in Section 8.1.1.2 of \citep{Westervelt2007}, RABBIT includes gear reducers between the motors and links that introduce significant joint friction and effectively eliminate passive joint motion. 
Such friction could be reintroduced later, using the same continuation process that we applied to slope and locomotion velocity.
As with the CGW, time was normalized by $t_\mr{o}=\sqrt{l_\mr{o}/g}$.
Given RABBIT's state~$\vec{x}=[\vec{q}~\dot{\vec{q}}]\T$ with configuration~$\vec{q}=[q_1~q_2~q_3~q_4~q_5]\T$, its continuous dynamics~$\vec{f}(\vec{x},\vec{u})$ and discrete dynamics~$\vec{g}(\vec{x})$ are derived via the Euler--Lagrange method\footnote{A symbolic derivation of the dynamics is available at \href{https://github.com/raffmax/IndirectMethod}{\url{github.com/raffmax/IndirectMethod}}}, as described in \cite{Westervelt2007}. In contrast to the CGW, the RABBIT's minimal coordinates are with respect to the inclined slope~$\gamma$, yielding the touch-down event
\begin{subequations}\label{eq:eqConRABBIT}
\begin{equation}\label{eq:eventRABBIT}
\begin{aligned}
    e=&~l_\text{t}\big(\cos(q_2 + q_4 + q_5) - \cos(q_1 + q_3 + q_5) \big) \\ &+l_\text{f}\big(\cos(q_2 + q_5) - \cos(q_1 + q_5)\big),
\end{aligned}
\end{equation}
where $l_\text{t}$ and $l_\text{f}$ denote the lengths of RABBIT's tibia and femur, respectively.
As before, the robot is required to walk at a prescribed average speed~$v_\text{avg}$, specified under a single operating condition ($n_\omega=1$):
\begin{equation}\label{eq:operatingRABBIT}
\begin{aligned}
    \omega=&~l_\text{t}\big( \sin(q_1 + q_3 + q_5)-\sin(q_2 + q_4 + q_5)\big) \\ &+ l_\text{f}\big(\sin(q_1 + q_5)-\sin(q_2 + q_5)\big) -v_\text{avg}T.
\end{aligned}
\end{equation}
\end{subequations}
Akin to the CGW optimization problem, the constraint~$\vec{h}\big(T,\vec{x}(T),\vec{x}(0);\vec{\sigma}\big)=\vec{0}$ of $\mathcal{P}_{\vec{\sigma}}$ follows from \eqref{eq:eventRABBIT} and~\eqref{eq:operatingRABBIT}, with $\vec{\sigma}^\mathrm{T}=[\gamma~v_\text{avg}]$.
As before, the objective was to minimize the cost of transport, defined as
\begin{equation}\label{eq:COTRabbit}
    c = \frac{y(T)}{mg\Delta x}\quad\Bigg\vert 
     \quad\dot{y} =\underbrace{\tfrac{k}{2}~\sum_{i=1}^4 u_i^2}_{=l(\vec{x},\vec{u})},~\Delta x = v_\text{avg}T,
\end{equation}
where~$k=(m\sqrt{g l_\mathrm{o}^3})^{-1}$ relates to the DC motor speed--torque gradient, assuming four identical motors with a $50\!:\!1$ gear ratio~\citep{Chevallereau2003}.

\subsection{From Passive to Level Ground Walking}\label{sec:passive2LevelRABBIT}

As in the CGW example, we again leveraged RABBIT’s passive optimal gaits to initialize Algorithm~\ref{algo:OptimalContinuation}, thereby generating optimal gait families within the parameterized problem~$\mathcal{P}_{\vec{\sigma}}$. Figure \ref{fig:subpassiveRABBIT} depicts the passive gait family that has been again systematically generated from a standstill as described in \citep{Rosa2023}. While there also exist passive gait families with longer periods originating from a standing configuration, these gaits include phases of knee hyperextension and exhibit a smaller region of convergence. Applying these longer-period gaits within Algorithm~\ref{algo:OptimalContinuation} therefore produces optimal solutions that are likely to be difficult to stabilize in practice.

We used the same error tolerances as before when evaluating $\vec{r}=\vec{0}$ and in the variable step-size integrator. To accelerate the evaluation of $\vec{r}$, we adopted the explicit input formulation described in \eqref{eq:CGWu}, transforming the DAE into an ODE that is solved using MATLAB's \texttt{ode45}. In particular, the initial input~$\vec{u}_0$ in the residual~\eqref{eq:r_ParamIndirect} is computed explicitly:
\begin{equation}\label{eq:RABBITu}
    \vec{u}_0 = -\frac{1}{kq}\frac{\partial \vec{f}}{\partial \vec{u}}\bigg\vert_{\vec{x}_0}\T \cdot \vec{p}_0.
\end{equation}
As with the CGW in Section~\ref{sec:passive2Level}, we generated a 1D family of optimal gaits on a varying slope by employing the scalar continuation parameter~$\sigma := \gamma$, while constraining the average speed to~$v_\text{avg} = 0.05 \sqrt{gl_\text{o}}$. 
We initialized Algorithm~\ref{algo:OptimalContinuation} with a passive gait which served as the optimal passive solution, represented by the tuple~$(T_\text{pas.}^\ast, \vec{x}_\text{pas.}^\ast(\cdot), \vec{u}_\text{pas.}^\ast(\cdot))$, with zero input~$\vec{u}_\text{pas.}^\ast(\cdot) \equiv \vec{0}$ and cost~$c = 0$ for the parameterized problem~$\mathcal{P}_{\gamma}$.

Again, for the cost of transport~\eqref{eq:COTRabbit}, costate and multiplier reconstruction was straightforward (Remark~\ref{rem:uSquared}). Thus, Algorithm~\ref{algo:OptimalContinuation} was initialized with
\begin{subequations}
\begin{align}
    \vec{\chi}_\text{start} &= \begin{bmatrix}
        T_\text{start} \\
        \vec{x}_{0,\text{start}} \\
        \vec{p}_{0,\text{start}} \\
        q_\text{start} \\
        \vec{u}_{0,\text{start}} \\
        \vec{\lambda}_\text{start}
    \end{bmatrix} = \begin{bmatrix}
        T_\text{pas.}^\ast \\
        \vec{x}_\text{pas.}^\ast(0) \\
        \vec{0} \\
        (mgv_\text{avg} T_\text{pas.}^\ast)^{-1} \\
        \vec{0} \\
        \vec{0}
    \end{bmatrix}, \\
    \sigma_\text{start} &= \gamma_\text{pas.},
\end{align}
\end{subequations}
where $v_\text{avg} = 0.05 \sqrt{gl_\text{o}}$ was fixed throughout the algorithm.
To generate a gait family that includes level-ground walking, the desired range of the slope parameter was set to $[\sigma_\text{start}, \sigma_\text{end}] = [\gamma_\text{pas.}, 0^\circ]$. 
Figure~\ref{fig:passive2activeRABBIT}a shows a projection of the resulting gait family $\vec{r}^{-1}(\vec{0})$ (red), generated by Algorithm~\ref{algo:OptimalContinuation}.
Figure~\ref{fig:passive2activeRABBIT}b shows the initial costates~$\vec{p}_0$, which vanish at the passive walking slope~$\gamma_\text{pas.}$, where $v_\text{avg}=0.05\sqrt{gl_\text{o}}$.

\begin{figure}[t]
    \centering
        \centering
    \begin{subfigure}{\columnwidth}
        \centering
        \includegraphics[width=0.99\linewidth]{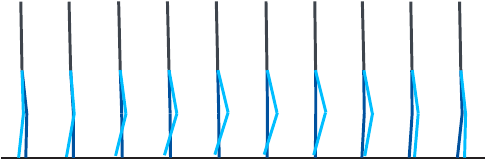}
    \end{subfigure}\\[4mm]
    \begin{subfigure}{\columnwidth}
        \centering
        \includegraphics[width=\linewidth]{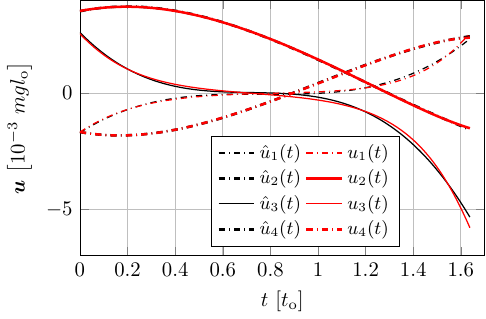}
    \end{subfigure}
    \caption{Optimal walking gait of RABBIT at an average speed of $v_\text{avg}=0.05\sqrt{gl_\text{o}}$ on level ground ($\gamma=0^\circ$). Keyframes show successive configurations of RABBIT throughout the gait cycle, while input trajectories compare indirect (red) and direct (black) shooting methods. The direct method parameterizes each input~$\hat{u}_i$ with a cubic B-spline ($n_\xi=4$).}
    \label{fig:visualGaitRABBIT}
\end{figure}

The final gait corresponds to level-ground walking ($\gamma=0^\circ$).
Figure~\ref{fig:visualGaitRABBIT} shows key frames of RABBIT's periodic motion and the optimal input trajectories from both indirect and direct shooting.
Using one cubic B-spline with $n_\xi=4$ per input~$\hat{u}_i$, direct shooting yielded a relative cost increase of $\frac{\hat{c}-c}{c}=0.0013$, comparable to the compass-gait walker in Figure~\ref{fig:directVSindirect}b.
\section{Conclusion \& Discussion}
\label{sec:disc}
In this work, we have provided advancements in periodic trajectory optimization for systems with hybrid dynamics, particularly in the context of legged locomotion.
Our key contributions include the formalization of a general trajectory optimization problem that extends first-order necessary conditions to a wider class of cost functions and periodic hybrid systems (Proposition~\ref{prop:firstOrderNecessary}). 
We introduced a framework for generating and analyzing families of optimal gaits using the indirect shooting method, coupled with numerical continuation techniques.
A major feature of this approach is the initialization of gait libraries with optimal periodic motions that exhibit vanishing actuation (passive gaits), for which Lagrange multipliers and costates can be effectively reconstructed (Lemma~\ref{lem:observe}, Proposition~\ref{prop:multipliers}). 
By comparing the indirect and direct shooting methods on the compass-gait walker, we demonstrated that the indirect approach achieves higher accuracy in generating optimal solutions while more effectively addressing the inherent challenges of input-space parameterization.
Furthermore, the successful application of indirect shooting with continuation methods (Algorithm~\ref{algo:OptimalContinuation}) to the high-dimensional RABBIT model demonstrates its applicability beyond simplified academic examples.

Despite these advancements, several important areas remain open for further research.
While our results suggest that the indirect shooting method offers faster function evaluations compared to the direct method in the context of the compass-gait walker, a more detailed analysis of the numerical complexity involved in the indirect method is still needed. A systematic comparison with direct collocation or multiple shooting frameworks would require carefully controlling for implementation-dependent factors such as solver configurations, discretization strategies, and integration schemes. Accounting for these benchmarking considerations will provide a clearer understanding of the computational trade-offs between indirect and direct approaches and will support broader conclusions about their relative efficiency as system complexity increases. Conducting such a study represents an important direction for future research, particularly for higher-dimensional legged and hybrid dynamical systems where scalability and numerical sensitivity are central concerns.

Furthermore, the first-order necessary conditions derived in this work are currently tailored to simple hybrid systems, characterized by a single continuous and a single discrete dynamic map.
A critical direction for future research involves extending these conditions to address more complex locomotion patterns, such as running or walking with a double support phase, where foot lift-off and touch-down do not occur simultaneously. 
Such extensions would require incorporating kinetic events (such as a foot leaving the ground when ground reaction forces reach zero) into the event definitions. 
Currently, our event functions are solely defined kinematically or by time, but this extension would necessitate incorporating actuation inputs~($\vec{u}$) into the definition of the event function. 
However, this extension is not straightforward, because when the event function explicitly depends on $\vec{u}$, the standard calculus of variations procedure used in Section \ref{sec:necessaryConditionsDerivation} requires $\vec{u}$ to be continuously differentiable. A less restrictive and more general alternative is to employ tools from Pontryagin's maximum principle.
Moreover, the introduction of multiple hybrid phases would convert the current two-point boundary problem into a multi-point boundary problem.
This transformation could be effectively addressed using a multiple shooting approach, where the initial shooting conditions align with the image of the discrete (reset) maps.

Another important future direction involves the integration of inequality path constraints into the indirect method. 
Addressing such constraints (e.g., joint limits or actuator torque limits) would require dividing the time domain into subarcs where these inequality constraints are either active or inactive \citep{Betts2010}.
This could also be addressed by introducing a multiple shooting scheme.
However, a major challenge here is the unknown number and order of these subarcs, which complicates the application of variable step-size shooting methods. 
Developing strategies to handle this complexity would significantly enhance the versatility of the indirect method and broaden its applicability to more complex, real-world systems where inequality constraints are critical.

While there is room for further refinement and extension of the indirect method for periodic trajectory optimization, this work underscores the advantages of the indirect approach over traditional direct methods, particularly in terms of accuracy.
The methodologies and insights presented in this paper mark a significant step forward in the application of indirect methods to the challenging and dynamic field of legged robotic systems.

\bibliographystyle{abbrvnat} %{plain} 
\bibliography{References}

\begin{ack}
This work was funded by the Deutsche Forschungsgemeinschaft (DFG, German Research Foundation) -- 501862165. It was further supported through the International Max Planck Research School for Intelligent Systems (IMPRS-IS) for Maximilian Raff.
\end{ack}

\appendix
%%%%%%%%%%%%%%%%%%%%%%%%%%%%%%%%%%%%%%%%%%%%%%%%%%
%%%%%%%%%%%%%%%%%%%%%%%%%%%%%%%%%%%%%%%%%%%%%%%%%%
% Variation                                          %
%%%%%%%%%%%%%%%%%%%%%%%%%%%%%%%%%%%%%%%%%%%%%%%%%%
\section{First-Order Variation}
\label{sec:appendixVariation}
\begin{figure*}[t]
%\vspace{9pt}
\hrule height 0.5pt % Top horizontal line
\centering
\begin{subequations}\label{eq:L1}
\begin{align}
    \frac{\mr{d} L_1}{\mr{d} \varepsilon}\bigg\vert_{\varepsilon=0}=&~\frac{\mr{d} c}{\mr{d} \varepsilon}\bigg(\tilde{T},\vec z^\ast(\tilde{T})+\varepsilon \cdot\delta \vec z(\tilde{T})\bigg) \bigg\vert_{\varepsilon=0}+\frac{\mr{d}}{\mr{d} \varepsilon}\bigg(\tilde{\vec{\lambda}}\T\vec h\big(\tilde{T},~\vec x^\ast(\tilde{T})+\varepsilon \delta \vec x(\tilde{T}),~\vec x^\ast(0)+\varepsilon  \delta \vec x(0)\big)\bigg)\bigg\vert_{\varepsilon=0}\\
    =&~\left[\frac{\partial c}{\partial T}\bigg\vert_\ast+\frac{\partial c}{\partial \vec z_T}\bigg\vert_\ast\dot{\vec{z}}^\ast{(T^\ast)}+{\vec{\lambda}^\ast}\T\left(\frac{\partial \vec{h}}{\partial T}\bigg\vert_\ast+\frac{\partial \vec{h}}{\partial \vec x_T}\bigg\vert_\ast\dot{\vec{x}}^\ast{(T^\ast)}\right)\right]\delta T+\delta \vec{\lambda}\T\vec h\vert_\ast\notag\\
    &~+\frac{\partial c}{\partial y_T}\bigg\vert_\ast\delta y(T^\ast)+\left[\frac{\partial c}{\partial \vec{x}_T}\bigg\vert_\ast+{\vec{\lambda}^\ast}\T\frac{\partial \vec{h}}{\partial \vec x_T}\bigg\vert_\ast\right]\delta \vec{x}(T^\ast)+{\vec{\lambda}^\ast}\T\frac{\partial \vec{h}}{\partial \vec x_0}\bigg\vert_\ast\delta \vec{x}(0),
\end{align}
\end{subequations}
\centering
\begin{subequations}\label{eq:L2}
\begin{align}
    \frac{\mr{d} L_2}{\mr{d} \varepsilon}\bigg\vert_{\varepsilon=0}=&~\frac{\mr{d}}{\mr{d} \varepsilon}\bigg(\tilde{\vec\lambda}_\mr{x}\T\vec{g}\big(\vec{x}^\ast(\tilde{T})+\varepsilon \delta \vec{x}(\tilde{T})\big)-\tilde{\vec\lambda}_\mr{x}\T\big(\vec{x}^\ast(0)+\varepsilon\delta \vec{x}(0)\big)+\tilde{\lambda}_\mr{y} \big(y^\ast(0)+\varepsilon\delta y(0)\big)\bigg)\bigg\vert_{\varepsilon=0}\\
    =&~{{}_\mr{x}\vec\lambda^\ast}\T\frac{\partial\vec{g}}{\partial \vec{x}_T}\bigg\vert_\ast\dot{\vec{x}}^\ast(T^\ast)\delta T+{{}_\mr{x}\vec\lambda^\ast}\T\frac{\partial\vec{g}}{\partial \vec{x}_T}\bigg\vert_\ast\delta \vec{x}(T^\ast)-{{}_\mr{x}\vec\lambda^\ast}\T\delta \vec{x}(0)+{{}_\mr{y}\lambda^\ast}\delta \vec{y}(0) \notag\\
    &~+\delta {}_\mr{x}\vec\lambda\T\big(\vec{g}(\vec{x}^\ast(T^\ast))-\vec{x}^\ast(0)\big)+\delta{}_\mr{y}\lambda y^\ast(0),
\end{align}
\end{subequations}
\vspace{-4mm}
\centering
\begin{subequations}\label{eq:L3}
\begin{align}
    \frac{\mr{d} L_3}{\mr{d} \varepsilon}\bigg\vert_{\varepsilon=0}=
    &~\frac{\mr{d}}{\mr{d} \varepsilon}\Bigg(\int_0^{T^\ast+\varepsilon\delta T} \mathcal{H}(\vec w^\ast+\varepsilon\delta \vec{w})-\left(\vec \rho^\ast+\varepsilon\delta\vec{\rho}\right)\T \left(\dot{\vec z}^\ast+\varepsilon\delta \dot{\vec{z}}\right)\mr{d}t\Bigg) \Bigg\vert_{\varepsilon=0}\notag\\
    =&~\int_0^{T^\ast+\varepsilon\delta T} \frac{\mr{d}}{\mr{d} \varepsilon}\bigg(\mathcal{H}(\vec w^\ast+\varepsilon\delta \vec{w})-\left(\vec \rho^\ast+\varepsilon\delta\vec{\rho}\right)\T \left(\dot{\vec z}^\ast+\varepsilon\delta \dot{\vec{z}}\right)\bigg)\mr{d}t \Bigg\vert_{\varepsilon=0} +\bigg[\mathcal{H}\big(\vec{w}_{T^\ast}^\ast\big)-{\vec{\rho}^\ast_{T^\ast}}\T{\dot{\vec{z}}^\ast_{T^\ast}}\bigg]\delta T\label{eq:leibniz}\\
    =&~\int_0^{T^\ast} \bigg(\frac{\partial \mathcal{H}}{\partial \vec{w}}\bigg\vert_\ast\delta \vec{w}-\delta{\vec{\rho}}\T\dot{\vec{z}}^\ast-{\vec{\rho}^\ast}\T\delta\dot{\vec{z}}\bigg)\mr{d}t+\bigg[\mathcal{H}\big(\vec{w}_{T^\ast}^\ast\big)-{\vec{\rho}^\ast_{T^\ast}}\T{\dot{\vec{z}}^\ast_{T^\ast}}\bigg]\delta T\notag\\
    =&~\bigg[-{\vec{\rho}^\ast}\T\delta \vec{z}\bigg]_0^{T^\ast}+\int_0^{T^\ast} \bigg(\frac{\partial \mathcal{H}}{\partial \vec{w}}\bigg\vert_\ast\delta \vec{w}-\delta{\vec{\rho}}\T\dot{\vec{z}}^\ast+\dot{\vec{\rho}}^{\ast\mathop{\mathrm{T}}}\delta\vec{z}\bigg)\mr{d}t+\bigg[\mathcal{H}\big(\vec{w}_{T^\ast}^\ast\big)-{\vec{\rho}^\ast_{T^\ast}}\T{\dot{\vec{z}}^\ast_{T^\ast}}\bigg]\delta T\label{eq:intByParts}\\
    =&~{\vec{\rho}_0^\ast}\T\delta \vec{z}(0)-{\vec{\rho}_{T^\ast}^\ast}\T\delta \vec{z}(T^\ast)+\bigg[\mathcal{H}\big(\vec{w}_{T^\ast}^\ast\big)-{\vec{\rho}^\ast_{T^\ast}}\T{\dot{\vec{z}}^\ast_{T^\ast}}\bigg]\delta T\\
    &~+\int_0^{T^\ast}\bigg( \left[\frac{\partial \mathcal{H}}{\partial \vec{z}}\bigg\vert_\ast+\dot{\vec{\rho}}^{\ast\mr{T}}\right]\delta \vec{z}+\frac{\partial \mathcal{H}}{\partial \vec{u}}\bigg\vert_\ast\delta \vec{u}+\left[\frac{\partial \mathcal{H}}{\partial \vec{\rho}}\bigg\vert_\ast-\dot{\vec{z}}^{\ast\mr{T}}\right]\delta \vec{\rho}\bigg)\mr{d}t.
\end{align}
\end{subequations}
\hrule height 0.5pt % Bottom horizontal line
\end{figure*}
To derive the stationary condition of the new problem~\eqref{eq:problemStatement}, we perturb its solution~$\big({}_\mr{z}\vec{\lambda}^\ast, \vec v^\ast,\vec w^\ast(\cdot)\big)$ with $\delta{}_\mr{z}\vec{\lambda}\T = \left[\delta {}_\mr{x}\vec{\lambda}\T~\delta{}_\mr{y}\lambda\right]$, $\delta\vec v\T = \left[\delta T~\delta\vec \lambda\T\right]$ and $\delta\vec w(\cdot)\T = \left[\delta\vec z(\cdot)\T~\delta\vec u(\cdot)\T~\delta\vec \rho(\cdot)\T\right]$ such that
\begin{equation}
\begin{aligned}
    {}_\mr{z}\tilde{\vec{\lambda}} &= {}_\mr{z}\vec{\lambda}^\ast+\varepsilon \delta {}_\mr{z}\vec{\lambda},\\
    \tilde{\vec v} &= \vec v^\ast+\varepsilon \delta \vec v,\\ 
    \tilde{\vec w}(\cdot) &= \vec w^\ast(\cdot)+\varepsilon \delta \vec w(\cdot),
\end{aligned}
\end{equation}
where $\varepsilon$ is infinitesimally small.
Note that $\vec w(t)$ is a function of time and is thus, coupled to the variation of the period time~$T$ in $\vec{v}$ at the end of the trajectory:
\begin{equation}
\begin{aligned}
    \tilde{\vec w}(\tilde{T}) &= \vec w^\ast(\tilde{T})+\varepsilon \delta \vec w(\tilde{T})\\
    &= \vec w^\ast(T^\ast+\varepsilon\delta T)+\varepsilon \delta \vec w(T^\ast+\varepsilon\delta T).  
\end{aligned}
\end{equation}
At a locally optimal point of problem~$\mathcal{P}$, the first variation $\delta L|_{\ast}$ of the Lagrangian~\eqref{eq:lagrangian} must vanish.
With $\tilde{L}(\varepsilon):=L({}_\mr{z}\tilde{\vec{\lambda}},\tilde{\vec v}, \tilde{\vec w}(\cdot))$, the stationary condition takes the form
\begin{equation}\label{eq:stationaryCondVar}
    \delta L|_{\ast}:=\frac{\partial \tilde{L}(\varepsilon)}{\partial \varepsilon}\bigg\vert_{\varepsilon=0}=0.
\end{equation}
The first-order variation of~\eqref{eq:stationaryCondVar} is carried out across the three components of the Lagrangian~\eqref{eq:lagrangian}.
This is done in equations~\eqref{eq:L1}--\eqref{eq:L3}, where we used the Leibniz integral rule in equation~\eqref{eq:leibniz} and integration by parts in equation~\eqref{eq:intByParts}.
Taking the sum of equations~\eqref{eq:L1}--\eqref{eq:L3} with additional grouping of terms , and omitting the superscript~$\ast$, yields the following expression:
\begin{subequations}
\begin{equation}\label{eq:stationaryEq1}
\begin{aligned}
    0 =&~\delta L \\
    =&\int_0^T\bigg( \left[\frac{\partial \mathcal{H}}{\partial \vec z}+\dot{\vec{\rho}}\T\right]\delta \vec x+\left[\frac{\partial \mathcal{H}}{\partial \vec\rho}-\dot{\vec{z}}\T\right]\delta y\bigg)\mr{d}t \\
    &~+\int_0^{T} \bigg(\left[\frac{\partial \mathcal{H}}{\partial \vec u}\right]\delta \vec u\bigg)\mr{d}t \\
    &~+\begin{bmatrix}
        \delta \vec z_0\T &  \delta \vec z_T\T  & \delta T & \delta \vec{\lambda}\T & \delta {}_\mr{z}\vec{\lambda}\T
    \end{bmatrix} \cdot {}_\mr{\delta}\tilde{\vec{r}},
\end{aligned}
\end{equation}
where
\begin{align}
    {}_\mr{\delta}\tilde{\vec{r}}\T&=\begin{bmatrix}
        {}_{\vec z_0}\tilde{\vec{r}}\T & {}_{\vec z_T}\tilde{\vec{r}}\T & {}_T\tilde{r} & \vec{h}\T & {}_\text{z}\vec{h}\T
    \end{bmatrix},\label{eq:resEq}\\
    {}_{\vec z_0}\tilde{\vec{r}}\T &= \begin{bmatrix}
        {}_{\vec x_0}\tilde{\vec{r}}\T & {}_{y_0}\tilde{r}
    \end{bmatrix}\\
    {}_{\vec z_T}\tilde{\vec{r}}\T &= \begin{bmatrix}
        {}_{\vec x_T}\tilde{\vec{r}}\T & {}_{y_T}\tilde{r}
    \end{bmatrix}\\
    {}_\text{z}\vec{h}\T & =\begin{bmatrix} \big(\vec{g}(\vec{x}_T)-\vec{x}_0\big)\T & y_0 \end{bmatrix},  
\end{align}
with
\begin{align}
    {}_{\vec x_0}\tilde{\vec{r}}=&~\frac{\partial \vec{h}}{\partial \vec x_0}\T \vec{\lambda}+{\vec p_0}-{}_\mr{x}\vec{\lambda},\label{eq:r_x0}\\
    {}_{y_0}\tilde{r}=&~q_0+{}_\mr{y}\lambda,\label{eq:r_y0}\\
    {}_{\vec x_T}\tilde{\vec{r}}=&~\frac{\partial\vec{g}}{\partial \vec{x}_T}\T{{}_\mr{x}\vec{\lambda}}-\vec{p}_T+\frac{\partial c}{\partial \vec{x}_T}\T+\frac{\partial \vec{h}}{\partial \vec x_T}\T{\vec{\lambda}},\label{eq:r_xT}\\
    {}_{y_T}\tilde{r}=&~\frac{\partial c}{\partial y_T}-q_T,\label{eq:r_yT}\\
    {}_T\tilde{r}=&~\mathcal{H}(\vec{{w}}_T)+ \frac{\partial c}{\partial T}+{\vec{\lambda}}\T\frac{\partial \vec{h}}{\partial T}\label{eq:r_T}\\
    &~+{}_{\vec x_T}\tilde{\vec{r}}\T \cdot\dot{\vec{x}}_T+{}_{y_T}\tilde{r}\cdot\dot{y}_T.\label{eq:r_T2}
\end{align}
\end{subequations}
Since the problem~\eqref{eq:problemStatement} is unconstrained, the first-order variation~\eqref{eq:stationaryEq1} must hold for arbitrary perturbations~$\delta {}_\mr{z}\vec{\lambda}$, $\delta \vec{v}$ and $\delta \vec{w}(\cdot)$.
Thus, applying the fundamental lemma of the calculus of variations yields 
\begin{equation}\label{eq:optimalityEq1}
    \begin{aligned}
    \dot{\vec \rho}(t) &= -\frac{\partial \mathcal{H}}{\partial \vec z}\bigg\vert_{\vec{w}(t)}\T \hspace{-0.5mm},\\
    \dot{\vec z}(t) &= \frac{\partial \mathcal{H}}{\partial \vec \rho}\bigg\vert_{\vec{w}(t)}\T\hspace{-0.5mm},\\
    \vec{0}&=\frac{\partial \mathcal{H}}{\partial \vec u}\bigg\vert_{\vec{w}(t)}\T,
    \end{aligned}
\end{equation}
for all $t\in[0,T]$ and requires the residual~\eqref{eq:resEq} to vanish:
\begin{align}\label{eq:resEq2}
    \begin{bmatrix}
        {}_{\vec x_0}\tilde{\vec{r}}\T & {}_{y_0}\tilde{r} & {}_{\vec x_T}\tilde{\vec{r}}\T & {}_{y_T}\tilde{r} & {}_T\tilde{r} & \vec{h}\T & {}_\text{z}\vec{h}\T
    \end{bmatrix}=\vec{0}.
\end{align}
While the differential~\eqref{eq:optimalityEq1} constitute the first part of the necessary conditions,~\eqref{eq:resEq2} contains so-called transversality (${}_i\tilde{\vec{r}}$) and boundary ($\vec{h}$, ${}_\text{z}\vec{h}$) conditions.
Notably, from~\eqref{eq:optimalityEq1} follows: $\dot{q}=-\nicefrac{\partial \mathcal{H}}{\partial y}=0$.
This results in a constant Lagrange multiplier~$q$ which is derived from the transversality condition~\eqref{eq:r_yT} as $q(t) \equiv q_0=q_T =\nicefrac{\partial c}{\partial y_T}$.
%
% \begin{remark}\label{rem:abnormal}
%     The constant costate~$q$ is commonly referred to as an \emph{abnormal multiplier}, since it may vanish ($q=0$) in degenerate instances of problem~$\mathcal{P}$ (see Chapter~4 of~\cite{Liberzon2011}). This situation arises, for example, in time-optimal problems where the terminal cost is given by $c = T$.
% \end{remark}
Likewise, the transversality conditions~\eqref{eq:r_x0} and \eqref{eq:r_y0} can be rewritten as
\begin{align}\label{eq:lamxy}
    {}_\mr{x}\vec{\lambda}=\frac{\partial \vec{h}}{\partial \vec x_0}\T \vec{\lambda}+{\vec p_0},\quad
    {}_\mr{y}\lambda =-q \,.
\end{align}
Hence, ${}_\mr{x}\vec{\lambda}$ can be eliminated from~\eqref{eq:r_xT}.
Finally, when the residuals~${}_{\vec{x}_T}\tilde{\vec{r}}=\vec{0}$ and ${}_{y_T}\tilde{r}=0$ vanish, the additive terms~\eqref{eq:r_T2} vanish, thus simplifying the final transversality condition~\eqref{eq:r_T}.
With these simplifications,~\eqref{eq:optimalityEq1} and \eqref{eq:resEq2} yield the first-order necessary conditions.

\section{Predictor-Corrector Continuation}
\label{sec:appendixContinuation}
The following provides the mathematical details of the pseudo-arclength continuation method summarized in Algorithm~\ref{algo:OptimalContinuation}.
For the one-dimensional parameterization ($n_\sigma=1$), we introduce the augmented variable~$\vec{\nu}\in\Real^{N+1}$ and write the residual function~\eqref{eq:r_ParamIndirect} and its Jacobian as
\begin{subequations}
    \begin{align}
    &\vec{\nu}\T = \begin{bmatrix}\vec{\chi}\T & \sigma\end{bmatrix},\\
    &{\vec{r}}:\Real^{N+1}\to\Real^{N},\\
        &{\vec{R}}:\Real^{N+1}\to\Real^{N\times (N+1)}.
    \end{align}
\end{subequations}
The solution set~$\vec{r}^{-1}(\vec{0})$ implicitly defines a curve of stationary periodic solutions, which we trace using a pseudo-arclength predictor-corrector method adapted from~\cite{Allgower2003}.
\begin{subequations}
This method consists of a forward Euler integration step of size~$h>0$ and direction~$d\in\{-1,1\}$, given by
\begin{align}\label{eq:predictor}
    \vec{\nu}_\text{pred} = \vec{\nu} +hd\vec{\tau},
\end{align}
where $\vec{\tau}$, referred to as the \emph{tangent induced by ${\vec{R}}(\vec{\nu})$}, is defined by

\begin{align}\label{eq:tangentvector}
    \vec{\tau}:=\left\{ \vec{\tau}\in\mathbb{R}^{N+1}\left\vert\begin{array}{cc}
        {\vec{R}}(\vec{\nu})\cdot\vec{\tau} & =\vec{0}, \\[1mm]
         \lVert\vec{\tau}\rVert_2 & = 1, \\[1mm]
         \text{det}\left(\begin{bmatrix}
             {\vec{R}}(\vec{\nu})\\\vec{\tau}\T
         \end{bmatrix}\right) & >0,
    \end{array}\right.\right.
\end{align}
and a subsequent corrector step that approximately solves the optimization problem:
\begin{align}\label{eq:corrector}
    \vec{\nu}_\text{corr} \approx \underset{\vec{\nu}\in{\vec{r}}^{-1}(\vec{0})}{\text{argmin}}\left\lVert\vec{\nu}-\vec{\nu}_\text{pred}\right\rVert_2,
\end{align}
utilizing Newton's method.
These steps are repeated until the desired family of gaits is constructed within a specified parameter range~$[\sigma_\text{start}, \sigma_\text{end}]$. 
An initial solution~$\vec{\chi}_\text{start}$ to the residual~\eqref{eq:r_ParamIndirect} at the parameter~$\sigma_\text{start}$ must be known or constructed from a passive optimal gait (Section~\ref{sec:reconstructLagrange}). 
Assuming that~$\sigma_\text{start}$ corresponds to a regular point $\vec{\nu}_\text{start}$ (i.e., the Jacobian ${\vec{R}}(\vec{\nu}_\text{start})$ has full rank), the tangent vector~\eqref{eq:tangentvector} is uniquely defined~\citep{Allgower2003}.
However, the corresponding curve~${\vec{r}}^{-1}(\vec{0})$ that passes through $\vec{\nu}_\text{start}$ can always be traced in two directions along its tangent vector~$\vec{\tau}_\text{start}$. 
To initiate the continuation process in the direction of the parameter~$\sigma_\text{end}$, we define
\begin{equation}\label{eq:Direction}
    d = \text{sign}\big(\sigma_\text{end}-\sigma_\text{start}\big)\cdot\text{sign}\big([0~\cdots~0~1]\cdot\vec{\tau}_\text{start}\big).
\end{equation}
\end{subequations}

%%%%%%%%%%%%%%%%%%%%%%%%%%%%%%%%%%%%%%%%%%%%%%%%%%
%%%%%%%%%%%%%%%%%%%%%%%%%%%%%%%%%%%%%%%%%%%%%%%%%%
% CGW                                            %
%%%%%%%%%%%%%%%%%%%%%%%%%%%%%%%%%%%%%%%%%%%%%%%%%%
\section{Compass-Gait Walker Dynamics}
\label{sec:cgw}
With the minimal coordinates~$\vec{q}=[\theta_\mr{sw}~\theta_\mr{st}]\T$, the auxiliary angle~$\alpha(\vec{q})=\theta_\mr{st}-\theta_\mr{sw}$ and the leg length~$l_\mr{o}=a+b$, we define the continuous dynamics as
\begin{subequations}
\begin{equation}
    \vec{M}(\vec{q})\ddot{\vec{q}}+\vec{C}(\vec{q},\dot{\vec{q}})\dot{\vec{q}}+\vec{G}(\vec{q})=\vec{B}u,
\end{equation}
where
\begin{align}
    \vec{M} &= \begin{bmatrix}
    m_\mr{h}b^2 & -m_\mr{l}l_\mr{o}b\cos(\alpha)\\-m_\mr{l}l_\mr{o}b\cos(\alpha) & (m_\mr{h}+m_\mr{l})l_\mr{o}^2+ma^2\end{bmatrix},\\
    \vec{C} &= \begin{bmatrix}
        0 & m_\mr{l}l_\mr{o}b\sin(\alpha)\dot{\theta}_\mr{st}\\ -m_\mr{l}l_\mr{o}b\sin(\alpha)\dot{\theta}_\mr{sw} & 0
    \end{bmatrix},\\
    \vec{G} &= \begin{bmatrix}
 m_\mr{l}bg\sin(\theta_\mr{sw})\\ -(m_\mr{h}l_\mr{o}+m_\mr{l}a+m_\mr{l}l_\mr{o})g\sin(\theta_\mr{st})
    \end{bmatrix},\\ 
    \vec{B}&=\begin{bmatrix}
        -1\\1
    \end{bmatrix}.
\end{align}
\end{subequations}
Hence, with the state~$\vec{x}\T=[\vec{q}\T~\dot{\vec{q}}\T]$, the vector field takes the form
\begin{equation}
    \vec{f}(\vec{x},u)=\begin{bmatrix}
        \dot{\vec{q}}\\\vec{M}(\vec{q})^{-1}\big(\vec{B}u-\vec{C}(\vec{q},\dot{\vec{q}})\dot{\vec{q}}-\vec{G}(\vec{q})\big)
    \end{bmatrix}.
\end{equation}
The equations for the collision of the swing-leg with the ground are derived by conservation of angular momentum:
\begin{subequations}
    \begin{equation}
        \vec{Q}^+(\vec{q})\dot{\vec{q}}^+=\vec{Q}^-(\vec{q})\dot{\vec{q}}^-,
    \end{equation}
    where
    \begin{align}
        \vec{Q}^- =&~\begin{bmatrix}
            0 & (m_\mr{h} l_\mr{o}^2+2m_\mr{l} al_\mr{o})\cos(\alpha)\\ 0 & 0
        \end{bmatrix}\notag \\
        &~-m_\mr{l}ab \begin{bmatrix}
            1 & 1\\ 0 & 1
        \end{bmatrix},\\
        \vec{Q}^+ =&~ \begin{bmatrix}
            m_\mr{l}b^2 & m_\mr{l}l_\mr{o}^2+m_\mr{l}a^2+m_\mr{h}l_\mr{o}^2\\ m_\mr{l} b^2 & 0
        \end{bmatrix}\notag \\
        &~-m_\mr{l}b l_\mr{o}\cos(\alpha)\begin{bmatrix}
            1 & 1\\ 0 & 1
        \end{bmatrix},
    \end{align}
\end{subequations}
where $\dot{\vec{q}}^-$ and $\dot{\vec{q}}^+$ are the velocities before and after impact, respectively.
Note that the relabeling of coordinates~$\vec{q}^+=[\theta_\mr{st}~\theta_\mr{sw}]\T$ extends the half stride to a complete periodic motion:
\begin{equation}
    \vec{g}(\vec{x}) = \begin{bmatrix}
        \begin{bmatrix}
            0 & 1\\ 1 & 0
        \end{bmatrix}\vec{q} \\
        \vec{Q}^+(\vec{q})^{-1}\vec{Q}^-(\vec{q})\dot{\vec{q}}
    \end{bmatrix}.
\end{equation}

\section{Direct-Method Implementation Details}
\label{sec:appendixDirect}

\subsection{Transcription of Input Space}\label{sec:inputTransc}
While we implement the Bézier polynomial in its explicit form:
\begin{equation}
    \hat{\vec{u}}(t,\vec{\xi})=\sum_{j=1}^{n_\xi}\begin{pmatrix}
        n_\xi-1\\j-1
    \end{pmatrix}(1-t)^{n_\xi-j}t^{j-1}{\xi}_{j},
\end{equation}
the cubic B-spline is defined by the recursion formula due to de~Boor, Cox, and Mansfield~[Chapter~2.2 in \cite{Piegl1997}]:
\begin{subequations}
\begin{alignat}{2}
    \mathcal{B}_{i,k}(t)&:=\left\{\begin{array}{ll}
        1 & \text{if~}\tau_i \leq t < \tau_{i+1}, \\
        0 & \text{otherwise},
    \end{array}\right.\quad &\text{if~}k=0\\
    \mathcal{B}_{i,k}(t)&:=\begin{array}{l}\frac{t-\tau_i}{\tau_{i+k}-\tau_i}\mathcal{B}_{i,k-1}(t)\\+\frac{\tau_{i+k+1}-t}{\tau_{i+k+1}-\tau_{i+1}}\mathcal{B}_{i+1,k-1}(t)\end{array},\quad &\text{if~}k>0
\end{alignat}
such that 
\begin{equation}
    \hat{\vec{u}}(t,\vec{\xi})=\sum_{j=1}^{n_\xi}\mathcal{B}_{j-1,n_\text{poly}}(t){\xi}_{j},
\end{equation}
where the polynomial degree~$n_\text{poly}=3$ corresponds to the cubic spline.
To define the knot points $\tau_i$, we uniformly separate the time by equidistant time steps $\Delta t=\nicefrac{T}{n_\text{seg}}$, where the number of segments is given by $n_\text{seg}=n_{\xi}-n_\text{poly}$.
Requiring the B-spline to be defined in the time interval~$[0,T]$ yields the knot points:
\begin{equation}
        \tau_i =
        (i-n_\text{poly})\Delta t, \quad \forall i\in\{0,1,\dots,n_\tau-1\}.
\end{equation}
where $n_\tau=2n_\text{poly}+n_\text{seg}$.
\end{subequations}
\subsection{Computation of Sensitivities}\label{sec:sens}
In addition to the numerical integration of $\hat{\vec{x}}(T)$ within the equality constraints~\eqref{eq:hHat}, the evaluation of the zero-function~$\hat{\vec{r}}$, as defined in equation~\eqref{eq:rHat}, depends on sensitivity information based on $\hat{\vec{s}}\T = [T~\vec{x}_0\T~\vec{\xi}\T]$.
There are three numerical approaches for computing the sensitivities $\nicefrac{\partial \hat{c}}{\partial \hat{\vec{s}}} \in \Real^{1 \times (1+n_{\text{x}}+n_{\xi})}$ and $\nicefrac{\partial \hat{\vec{h}}}{\partial \hat{\vec{s}}} \in \Real^{(n_{\text{x}}+2) \times (1+n_{\text{x}}+n_{\xi})}$: finite differences, forward sensitivities and adjoint techniques~\citep{Sengupta2014}.
Generally, finite difference methods require extensive manual tuning of step sizes to achieve the desired accuracy in sensitivity information. In contrast, when using variable step-size solvers, this accuracy is automatically maintained within the solver's tolerances for both forward sensitivities and adjoint techniques.
We chose to implement forward sensitivities, which compute the sensitivity of the flow $\nicefrac{\partial \hat{\vec{z}}}{\partial \hat{\vec{s}}}$ by simultaneously integrating the dynamics $\dot{\vec{z}}\T = [\vec{f}\T~l]$ and their derivatives, as described in~\cite{Dickinson1976}.
While forward sensitivities are straightforward to implement, their complexity scales as $\mathcal{O}(n_\text{x}n_\xi)$, which can become computationally expensive when the input parameterization is large ($n_\xi \gg n_\text{x}$).
In such cases, the adjoint method is more efficient, with a complexity of $\mathcal{O}(n_\text{x} + n_\xi)$, as it only provides sensitivity information at the final time~$T$. 
Since only the sensitivities at the final time~$T$ are needed to compute $\hat{\vec{r}}$, the adjoint method avoids the overhead of forward sensitivity methods, which return sensitivities at all times~$t\in[0,T]$.
\subsection{Second Order Condition}\label{sec:2ndCond}
To determine whether a zero~$\hat{\vec{\chi}} \in \hat{\bar{\vec{r}}}^{-1}(\vec{0})$ corresponds to a local minimum or a saddle point, we can examine the matrix~$\hat{\bar{\vec{R}}}$, which is obtained as a by-product of applying Newton's Method in Algorithm~\ref{algo:OptimalContinuation}.
Introducing the gradient notations $\nabla \hat{c}\T = \frac{\partial \hat{c}}{\partial \hat{\vec{s}}}$ and $\nabla \hat{\vec{h}}\T = \frac{\partial \hat{\vec{h}}}{\partial \hat{\vec{s}}}$, the Jacobian matrix takes the specific form:
\begin{equation}
    \hat{\bar{\vec{R}}} = \begin{bmatrix}
        \nabla^2 \hat{c} + \nabla^2 \hat{\vec{h}} \hat{\vec{\lambda}} & \hat{\vec{h}} \\
        \hat{\vec{h}}\T & \vec{0}
    \end{bmatrix}.
\end{equation}
Assuming that the equality constraints~$\hat{\vec{h}}$ are regular (i.e., $\nabla \hat{\vec{h}}$ has full rank), the second-order condition for a local minimum requires the projected Hessian, defined as
\begin{equation}
\begin{aligned}
    \hat{\vec{H}} &= \left(\begin{bmatrix}
        \vec{I} \\ \vec{0}
    \end{bmatrix} \hat{\vec{D}}\right)\T \hat{\bar{\vec{R}}} \left(\begin{bmatrix}
        \vec{I} \\ \vec{0}
    \end{bmatrix} \hat{\vec{D}}\right) \\
    &= \hat{\vec{D}}\T \left( \nabla^2 \hat{c} + \nabla^2 \hat{\vec{h}} \hat{\vec{\lambda}} \right) \hat{\vec{D}},
\end{aligned}
\end{equation}
to be positive semi-definite for any~$\hat{\vec{D}} \in \mathbb{R}^{n_\text{x}+n_\xi+1 \times n_\xi-1}$ that satisfies $\text{span}(\hat{\vec{D}}) = \text{null}(\nabla \hat{\vec{h}}\T)$ \citep{Luenberger2021}.

In practice, the matrix~$\hat{\vec{D}}$, containing linearly independent columns, can be constructed using the singular value decomposition (SVD) of~$\nabla \hat{\vec{h}}$. Hence, for a zero~$\hat{\vec{\chi}} \in \hat{\bar{\vec{r}}}^{-1}(\vec{0})$, checking the sign of the smallest eigenvalue of~$\hat{\vec{H}}$ is sufficient to distinguish between a local minimum and a saddle point.
This classification can be summarized compactly as:
\begin{equation}
    \min\big(\text{eig}(\hat{\vec{H}})\big) \quad \left\{
    \begin{array}{ll}
        < 0, & \text{saddle point}, \\
        = 0, & \text{local minimum}, \\
        > 0, & \text{strict local minimum}.
    \end{array}\right.
\end{equation}

\end{document}